\documentclass[11pt]{article}
\usepackage{fullpage}
\usepackage{graphicx}
\usepackage{amsthm}
\usepackage{amsmath}
\usepackage{amsfonts}
\usepackage{caption}
\usepackage{subcaption}
\usepackage{natbib}
\usepackage[boxed]{algorithm2e}

\begin{document}

\author{
  Jonathan Rosenski\\
  Weizmann Institute of Science \\
  \texttt{jonathan.rosenski@weizmann.ac.il}
  \and
  Ohad Shamir\\
  Weizmann Institute of Science \\
  \texttt{ohad.shamir@weizmann.ac.il}
  \and
  Liran Szlak\\
  Weizmann Institute of Science \\
  \texttt{liran.szlak@weizmann.ac.il}
}

\date{}

\title{Multi-Player Bandits -- a Musical Chairs Approach}
\maketitle

%Conventions: Use users not players, arms not machines, static setting not fixed setting, MEGA not mega.

\begin{abstract}
We consider a variant of the stochastic multi-armed bandit problem, where multiple players simultaneously choose from the same set of arms and may collide, receiving no reward. This setting has been motivated by problems arising in cognitive radio networks, and is especially challenging under the realistic assumption that communication between players is limited. We provide a communication-free algorithm (Musical Chairs) which attains constant regret with high probability, as well as a sublinear-regret, communication-free algorithm (Dynamic Musical Chairs) for the more difficult setting of players dynamically entering and leaving throughout the game. Moreover, both algorithms do not require prior knowledge of the number of players. To the best of our knowledge, these are the first communication-free algorithms with these types of formal guarantees. We also rigorously compare our algorithms to previous works, and complement our theoretical findings with experiments.
\end{abstract}

\section{Introduction}

The stochastic multi-armed bandit (MAB) problem is a classic and well-studied setting of sequential decision-making, which exemplifies the dilemma of exploration vs. exploitation (see \cite{bubeck2012regret} for a comprehensive review). In this problem, a player sequentially chooses from a set of actions, denoted as `arms'. At every round, each arm produces a reward sampled from some unknown distribution in $[0,1]$, and the player receives that reward, but does not observe the reward of other arms. The player's goal, of course, is to maximize the cumulative reward. The dilemma of exploration vs. exploitation here is that the more the player `explores' by trying different arms, she will have a better understanding of each machine's expected reward. The more the player `exploits' the machine which she thinks is best, the less rounds are wasted on exploring bad machines. 

In this work, we study a variant of this problem, where there are many players who choose from the same set of arms. If two or more choose the same arm then there is a `collision' and no reward is provided by that arm. Moreover, we assume that players may not communicate. The goal is to find a distributed algorithm for players that will maximize the sum of their rewards. One motivation for this setting (as discussed below in the Related Work section below) comes from the field of cognitive radio networks, where several users utilize the same set of channels, in a situation where the quality of the different channels varies, and direct coordination between the players is not possible. We use the standard notion of (expected) regret to measure our performance, namely the difference between the expected cumulative reward of the arm with highest mean reward, and the expected cumulative rewards of the players. 

We focus on a particularly challenging situation, where the players cannot communicate, there is no central control, and the players cannot even know how many other players are also participating. At every round each player decides which arm to sample. After the round is over the player receives the reward associated with the chosen arm, or an indication that the arm was chosen by at least one other player, in which case they receive no reward. The event that more than one player chooses the same arm will be referred to as a collision.

We will consider two variants in this work - a \emph{static} setting in which all players start the game simultaneously and play for T rounds, and a \emph{dynamic} setting, in which players may enter and exit throughout the game. Our main results are the following:
\begin{itemize}
\item For the static case we propose and analyze the Musical Chairs (MC) algorithm, which achieves, with high probability and assuming a fixed gap between the mean rewards, a constant regret independent of $T$.
\item For the dynamic setting we propose the Dynamic Musical Chairs (DMC) algorithm, which achieves an $\tilde{O}(\sqrt{xT})$ regret (with high probability and assuming a fixed gap between rewards), where $x$ is a bound on the total number of players exiting and leaving.
\item We study the behavior of previous algorithms for this problem, and show that in the dynamic setting, there are some reasonable scenarios leading to their regret being linear in T. For other scenarios, we show that our regret guarantees improve on previous ones.
\item We present several experiments which validate our theoretical findings.
\end{itemize}
All guarantees hold assuming all players implement the algorithm, but do not require any communication or coordination during the game.

The paper is organized as follows: Section \ref{sec:setting} provides a formal description of the problem setting. Section \ref{sec:mc} introduces the algorithms and regret analysis for the static setting and dynamic setting. Section \ref{sec:mega} considers previous work, and studies scenarios where the regret performance substantially differ compared to our results. Section \ref{sec:experiments} presents some experiments. Section \ref{sec:discussion} provides concluding remarks, discussion, and open questions. Finally, Appendix \ref{app:proofs} contains most of the proofs.

\subsection*{Related Work}

Most previous work on multi-player multi-armed bandits assumed that players can communicate, and included elements such as a negotiation phase or exact knowledge of the number of players, which remains fixed throughout the game, e.g. \cite{liu2009distributed,anandkumar2011distributed, kalathil2014decentralized}. However, in modeling problems such as cognitive radio networks, where players may be unable or unwilling to coordinate, these are not always realistic assumptions. For example, the algorithm proposed in  \cite{liu2009distributed} relies on the players agreeing on a time division schedule for sharing the best arms. Each player stays on one of the best arms for a certain time period, and at the end of the time period, players switch. The algorithm requires all players to know the number of players, which needs to be fixed.\cite{anandkumar2011distributed} provide an algorithm which is communication and cooperation free, but requires knowledge of the number of players. In order to overcome this added requirement the authors present an algorithm which estimates the number of players. The main idea of the algorithm is to estimate the number of players based on the number of collisions that were seen and call this estimate $\hat{N}$, choose one of the best $\hat{N}$ arms uniformly at random, stay on that arm until a collision occurs, at which time the player repeats the procedure. The performance guarantees of this algorithm are rather vague, and do not hold for the dynamic setting. Rather than estimate the number of players directly, as the algorithm presented in this work does, their algorithm has an estimation technique which converges to the correct number of players, given that the number of players is static. Another approach which requires communication is the algorithm proposed in \cite{kalathil2014decentralized} in which players negotiate in order to reach an an agreement where every player chooses their own unique arm, and thus there are no collisions. Their algorithm, called $\mbox{dUCB}_4$, uses Bertsekas' auction algorithm to have each player choose their own unique arm. The paper also proposes an algorithm which address stochastic rewards changing according to a Markov process. This algorithm does not apply to our setting in which communication is forbidden. 
 
The work most similar to ours is \cite{avner2014concurrent} , where communication is not allowed and there is no knowledge of the number of players. The proposed algorithm, named MEGA, is based on an elegant combination of the well-known $\epsilon-$ greedy MAB algorithm with a collision avoidance protocol, known as the ALOHA protocol, used in signal and control processing. The $\epsilon-$greedy algorithm  (see e.g. \cite{sutton1998reinforcement}) is an algorithm for the setting of a single player in the multi armed bandit problem, which ensures that the majority of exploration occurs at the beginning of the game, and after accumulating sufficient information on the mean rewards,  most of the remaining iterations are used to exploit the arm with the highest expected profit. This is done by having an exploration probability $\epsilon_t$ that decreases as $\frac{1}{t}$ where $t$ is the current iteration. In every iteration, the player chooses with probability $\epsilon_t$ an arm uniformly at random, and with probability $1-\epsilon_t$ she exploits by choosing the arm with the highest empirical mean reward. The ALOHA protocol is a collision avoidance protocol used in multi-player signal processing schemes. The protocol dictates that a player, in the event of a collision, should decide, by some random process, whether to persist on the same arm, or to leave the arm and not return for a time period, chosen also at random. This time period is called the `unavailability time'.

\cite{avner2014concurrent} analyze the MEGA algorithm, and show that in the static setting, assuming parameters are chosen appropriately, it achieves $O\left(T^\frac{2}{3}\right)$ regret. A full analysis of the algorithm in the dynamic setting is not provided, but it is shown that there exist scenarios in the dynamic setting which will have an expected regret of $O\left( T^{\frac{2}{3}} \right)$. However, as we discuss in detail in section \ref{sec:mega}, the MEGA algorithm may perform poorly in some reasonable dynamic scenarios. Essentially, this is because collision frequency decreases as the game proceeds, but never reaches zero. Although the frequency can be tuned based on the algorithm's parameters, it is difficult to find a single combination of parameters that will work well in all scenarios. %In contrast, a principal idea of our algorithm is to allow a relatively short learning phase, where collisions are permitted, and then to fix the players' choice of arms once learning is concluded, thus preventing future collisions as well as achieving a configuration which is optimal with high probability.

%In the MEGA algorithm, collision frequency decreases as the game proceeds, but never reaches zero. However, in the MC and DMC algorithms that we propose, collisions happen only in a small subset of the rounds (the learning phase), and no collisions occur during the rest of the game. A principal idea of our algorithm is to fix the players' choice of arms once learning is concluded, thus achieving an optimal configuration with no collisions. 

\section{Setting}\label{sec:setting}

In the standard (single-player) stochastic MAB setting there are $K$ arms, with the rewards of each
arm $i\in [K] $ sampled independently from some distribution on $ [0,1] $, with expected
reward $ \mu_i$. Every round, a player chooses an arm and would like to receive the highest cumulative reward possibly in $T $ rounds overall. In this work, we focus for simplicity on the finite-horizon case, where $T$ is fixed and known in advance.

The multi-player MAB setting is similar, but with several players instead of a single one. In fact, we consider two cases: one where the set of players,  and therefore number of players $ N $, is fixed and another where the number of players, $ N_t $, can change at any round $ t $. In our model we would like to minimize, or even eliminate, any central control and communication, and assume that players do not even possess knowledge of the value of $N_t$. Generally, we assume $K$, $N$ ,and $N_t$ are all much smaller than $T$. We will denote by "the top $N$ arms" the set of $N$ arms with the highest expected rewards.

The performance in the standard single-player MAB setting is usually measured by how small is the regret (where we take expectations over the rewards of the arms):
\begin{gather*}
R := T\cdot \mu_{\ast} - \sum_{t=1}^T \mu\left(t \right)
\end{gather*}
where $\mu\left(t \right) $ is the expected reward of the arm chosen by the
single player at round $i$, and $\mu_{\ast} = \max_{i} \mu_i$ is the expected
reward of the arm with the highest expected reward. The regret is non-trivial
if it is sub-linear in $T$.

In the multi-player setting, we generalize this notion, and define our regret with respect to the best static allocation of players to arms (in expectation over the rewards), as follows:
\begin{gather*}
R := \sum_{t=1}^T \sum_{k\in K^{\ast}_t} \mu_k\left(t \right)  - \sum_{t=1}^T \sum_{j=1}^N \mu_j\left(t \right)  \cdot \left( 1 - \eta_j\left(t \right)\right)
\end{gather*}
where $\mu_j\left(t \right) $ is the expected reward of the arm chosen by player $j$ at round $t$, $ N_t $ is the number of players at round t, $K^{\ast}_t$ is the set of the highest $ N_t $ ranked arms where the rank is taken over the expected rewards, and $ \eta_j\left(t \right)  $ is a \emph{collision} indicator, which equals $ 1 $ if player j had a collision at round t, and $ 0 $ otherwise. We define a collision as the event where more than one player chose the same arm at a given round, and assume that no reward is obtained in that case.

Since achieving sublinear regret is trivially impossible when there are more players than arms, we assume throughout that the number of players, in both the static and dynamic settings, is always less than the number of arms. 

\section{Algorithms and Analysis}\label{sec:mc}

\subsection{The Musical Chairs (MC) Algorithm}
We begin by considering the static case, where no players enter or leave. The MC algorithm, that we present below for this setting, is based on the idea that after a finite time of random exploration, all players learned a correct ranking of all the arms with high probability (assuming gaps between the mean rewards). If after this time all players could fix on one of the top $N$ arms and never leave, then from this point onward, there would be no regret accumulating. The algorithm we present is composed of a learning phase, with enough rounds of random exploration for all players to learn the ranking of the arms and the number of players; a `Musical Chairs' phase, in which the $N$ players fix on the top $N$ arms; and a `fixed' phase where all players remain fixed on their arm.

\begin{algorithm}[h]
\TitleOfAlgo{MC}
\SetKwData{Left}{left}\SetKwData{This}{this}\SetKwData{Up}{up}
\SetKwFunction{Union}{Union}\SetKwFunction{FindCompress}{FindCompress}
\SetKwInOut{Input}{input}\SetKwInOut{Output}{output}
\SetKw{qq}{loop}
\Input{Parameters $ T_0 $, $ T_1 $}
\BlankLine
$ C_{T_0} = 0$, $\tilde{\mu}_i \left(t \right)  = 0 $, $ o_i = 0 $, $ s_i\left(0 \right)  = 0 $ $ \forall i\in {1,...,K} $\\
\For{t = 0 to $ T_0 $}{
sample arm $i \sim $ $ U\left({1,...,K} \right)  $ \\receive $ \eta \left(t \right)  $ and $ r\left(t \right)  $\\
\uIf{$ \eta \left(t \right)  \neq 1 $} {
update $ o_i = o_i + 1 $\\
$ s_i\left(t \right)  = s_i\left(t-1 \right)  + r\left(t \right)  $\\}
\Else{
$ C_{T_0} = C_{T_0} + 1 $
}
 }
$ \forall i \in {1,...,K} $ set $ \tilde{\mu}_i = \frac{s_i \left(T_0 \right) }{o_i} $\\
sort indices  in $[K]$ according to empirical mean in an array. Call this array $A$.\\
$ N^\ast := \mbox{round}\left(\frac{\log\left(\frac{T_0 - C_{T_0}}{T_0} \right) }{\log\left(1-\frac{1}{K} \right)  } + 1 \right)$ and $ N^\ast := K $ if $C_{T_0} = T_0 $\\
$ j= $ Musical Chairs($ N^\ast $ , A ) \\
stay fixed on arm $ j $ for the remainder of the rounds (total rounds is $ T_1 $ ) \\
 \textbf{Note: } \\
$ \eta $ is a collision indicator, $ N^\ast $ is the estimated number of players, $C_{T_0} $ is the number of collisions the player has experienced, $ \tilde{\mu}_i $ is empirical mean of rewards for arm $ i $, $ o_i $ is the number of successful observations of arm $ i $, and $ s_i $ is the sum of rewards received from arm $ i $ \\
\end{algorithm}\DecMargin{1em}

\begin{algorithm}[h!]
	\SetAlgorithmName{Subroutine}{subroutine}{Musical Chairs}
	\TitleOfAlgo{Musical Chairs}
	\SetKwInOut{Input}{input}
	\Input{Parameter $ N^\ast $, sorted array of arms $ A $ }
	\BlankLine
	\SetKw{qq}{loop for $T_1-T_0$ iterations}
	\qq$ \{ $
	\\sample $ i \sim U\left({1,...,N^\ast} \right)  $ and choose arm $ A[i] $\\
	receive $ \eta\left(t \right)  $\\
	\If{$ \eta\left(t \right)  == 0 $}{
		output $ A[i] $ and $ return $
	}$ \} $
\end{algorithm}

The Musical Chairs subroutine works by having each player randomly choose an arm in the top $N$ arms, until she chooses one without experiencing a collision. From that point onwards, she chooses only that arm. It can be shown that if all players implement this subroutine, then after a bounded number of rounds (in expectation), all players will fix on different arms, and there will be no more added regret.  The Musical Chairs subroutine's success depends on each player being able to accurately estimate a correct ranking of the machines (the ranking needs to be accurate enough to distinguish the best $N$ machines from the rest) and to estimate the correct value of $N$.

\subsection{Analysis of the MC algorithm}
Let $ N $ be the number of players and let $ i_1, ..., i_N $ denote the best $ N $ ranked arms. For each player we denote by $\tilde{\mu_j} $ to be that player's measured empirical mean reward of arm $j$. We use the following definition from \cite{avner2014concurrent}:
\theoremstyle{definition}
\newtheorem{defn}{Definition}
\begin{defn}
An $ \epsilon $-correct ranking of K arms is a sorted list of empirical mean rewards of arms such that $\forall i,j$ : $\tilde{\mu_i}$ is listed before $ \tilde{\mu_j} $ if $ \mu_i - \mu_j > \epsilon $
\end{defn}
\theoremstyle{plain}
\newtheorem{theorem}{Theorem}
\newtheorem{lemma}{Lemma}

\begin{theorem}
Let $\Delta>0$ be the gap between the expected reward of the $N$th best arm
and the $N+1$ best arm. Then for all  $\epsilon<\Delta$ and $\delta\in (0,1)$,  with probability $ \geq 1 - \delta $, the expected regret of $ N $ players using the MC algorithm with $ K $  arms for $ T $ rounds, with parameter $T_0$ set to\\
$ T_0 = \left\lceil \max\left( \frac{K}{2} \cdot \ln \left( \frac{2 \cdot K^2}{\delta} \right), \frac{16 \cdot K}{\epsilon^2} \cdot \ln\left( \frac{4 \cdot K^2}{\delta} \right)  , \frac{K^2 \cdot \log\left(\frac{2}{ \delta_2}  \right) }{ 0.02 }  \right) \right\rceil$ is at most
\[
T_0 \cdot N + 2 \cdot \exp(2) \cdot N^2 .
\]
  \label{thm:MC}
\end{theorem}
Note that the bound we give is in expectation over the rewards and the algorithm's randomness,  conditioned on the event (occurring with probability at least $1-\delta$) that players learn an $\epsilon$-correct ranking and estimate the true number of players.

The proof of theorem~\ref{thm:MC} is composed of three lemmas presented below, whose formal proof appears in appendix~\ref{app:proofs}.

We begin by showing that with high probability, all players will learn an $\epsilon$-correct ranking after a time period independent of $T$:
\begin{lemma}
$ \forall \epsilon > 0 $ and $ 0<\delta <1 $, then after $ T_0 = \left\lceil \max \left\{ \frac{K}{2} \cdot \ln \left( \frac{2 \cdot K^2}{\delta} \right), \frac{16 \cdot K}{\epsilon^2} \cdot \ln\left( \frac{4 \cdot K^2}{\delta} \right) \right\} \right\rceil$ rounds of random exploration, all players have an $ \epsilon- $correct ranking of the arms w.p. $ \geq 1 - \delta $
\label{lemma:mclearn}
\end{lemma}

We then show how estimating the number of players also requires a number of rounds independent of $T$ with high probability. Knowing the value of $N$ exactly is required in order for the players to run the Musical Chairs subroutine and choose an arm  from the top $N$ arms. To estimate $N$, each player keeps track of the number of collisions till time $t$, denoted as $C_t$, and after $T_0$ rounds, computes the estimate $ N^\ast = \mbox{round}\left(\log\left(\frac{T_0-C_{T_0}}{T_0} \right) /\log\left(1-\frac{1}{K} \right)   + 1 \right)$, where $\text{round}(\cdot)$ rounds to the nearest integer. The following lemma proves that $N^\ast$ will indeed equal $N$ with high probability:
\begin{lemma}
Let $ \delta \in [0,1] $. For $ \epsilon_1 = \frac{0.1}{K} $, if the number of rounds $T_0$ used to estimate $N$ is at least $\left\lceil \frac{\log\left(\frac{2}{ \delta}  \right) }{2{\epsilon_1}^2 } \right\rceil $, then w.p. $ \geq 1 - \delta $ we have that $ N^{\ast} = N $.
\label{lemma:unknown}
\end{lemma}

Finally, given that players were able to learn an $ \epsilon $-correct ranking and the number of players, we can upper bound the expected time (and hence the regret) for all the players to fix on different arms:
\begin{lemma}
Denote by $ R^F $ the regret accumulated due to players running the musical chairs subroutine. Conditioned on the event that all $N$ players learned an $\epsilon$-correct ranking and that $N^{\ast}=N$, it holds that the expected value of $R^F$ is at most $2 \cdot \exp(2) \cdot N^2 $.
\label{lemma:fixing}
\end{lemma}

%By setting the number of learning rounds to be big enough to learn a correct ranking and estimating the number of players, and then fixing on a machine in $K^\ast$, then there is no more build up of regret. Therefore by 
Combining the three lemmas above, we get Theorem~\ref{thm:MC}.

\subsection{The Dynamic Musical Chairs (DMC) Algorithm}

In this subsection we will consider the case when players can enter and leave.
For the dynamic setting we suggest an extension of the MC algorithm, which simply runs the algorithm in epochs and restarts at the end of each epoch (see pseudocode below). We call this algorithm the Dynamic MC (DMC) algorithm and it requires the use of a shared clock between all players, to synchronize the epochs epochs. We note that having a shared clock is a mild assumption which has been used previously in several works (See for example
\cite{avner2015learning}, \cite{shukla2014synchronization}, \cite{nieminen2009time}). This clock means that at any round $t$, players know what is $t \mod T_1$, where $T_1$ (a parameter of the algorithm) is the number of rounds in an epoch. However, communication between players is still not allowed, and the shared clock is not used for resource allocation or synchronization between players regarding which arm to choose.

We emphasize that in the dynamic setting, some restriction on the frequency at which players enter or leave is necessary for any algorithm to obtain a sub-linear regret bound. This is because if players may enter or leave at every round, then it is possible that no player stays long enough to even learn the
true ranking of any arm, in which case any algorithm will result in linear regret. For this reason, we assume that the overall number of players entering and leaving is sublinear in $T$. Moreover, since time periods are synchronized, we will
allow ourselves to assume that players can only enter and leave after the learning period in each epoch. We note that according to our analysis, the proportion of rounds belonging to learning period is a vanishing portion of the total number of rounds $T$, and therefore this assumption is not overly restrictive. Moreover, under some conditions, this assumption can be weakened to cover only leaving players, without significantly changing our regret bounds\footnote{For example, if the entering players can refrain from picking arms during the learning phase, instead accumulating regret. Since the total length of the learning phase is less than our regret bounds, this won't affect the bounds by more than a small constant.}.

\begin{algorithm}
\TitleOfAlgo{Dynamic MC}
\SetKwInOut{Input}{input}\SetKwInOut{Output}{output}
\SetKw{Kw}{loop}
\Input{Parameters $ T_0 , T_1, T $, current round $ t $}
\BlankLine
explore/learn until $ t\bmod{T_1} = 0 $\\
\Kw{ $ \{ $ }{\\
run MC ($ T_0 $, $ T_1 $)
$ \} $}
\end{algorithm}

\subsection{Analysis of DMC Algorithm}
%We will denote by $ X $ as the rate at which the event of a player leaving or entering happens, i.e. every $ X $ rounds a player either leaves or enters (or both, or none).

%Let $ X = T^\alpha $ for some $ 0 < \alpha < 1 $.
The main result here is the following theorem:
\begin{theorem}
Let $N_m\leq K$ be an upper bound on the number of active players at any time point;
$\Delta_{min}=\min_{i=1,\ldots,N_m}\mu_i-\mu_{i+1}$ the minimal gap between
the best $N_m+1$ arms, with a known lower bound $\epsilon>0$; and $x$ be an
upper bound on the total number of players entering and leaving during $T$ rounds.
Then with arbitrarily high probability, the expected regret of the DMC
algorithm (over the rewards), using $\Theta(\sqrt{xT})$ epochs with
$\tilde{O}(1)$ learning rounds at the beginning of each epoch, is at most:
\begin{gather*}
\tilde{O} \left( \sqrt{xT}\right)
\end{gather*}
where the $ \tilde{O}  $ hides factors logarithmic in $T,\delta$, and polynomial in $\Delta_{min},K,N_m$. 
\label{thm:DMC}
\end{theorem}
As in theorem~\ref{thm:MC}, the bound is in expectation over rewards and the algorithm's randomness, conditioned on the high-probability event that in each epoch, the players learn the correct ranking and the number of players. 

The bound in the lemma hides several factors to simplify the presentation. More specifically, the bound is based on the following lemma, and taking $T_1=\left \lceil \sqrt{ \frac{T \cdot \left(T_0 + 2\cdot T_f \right)}{ x  } } \right \rceil$:

%
%
%
%We will denote by $ e $ and $l$ the total number of players entering and leaving throughout the game, respectively.
%$ T_f \leq N \cdot e^2 $ be the expected time it takes any player to fix on an arm during the `Musical Chairs' period, and $ N_m $ be the maximum number of active players at any given round.
%Theorem~\ref{thm:DMC} is a result of the following lemma, and taking an
%appropriate epoch length parameter $T_1$:

%$\left \lceil \sqrt{ \frac{T \cdot N_m \cdot \left(T_0 + T_f \right)}{e + e
%\cdot \frac{N_m}{K} + l} } \right \rceil$, which can be lower bounded by
%$\left \lceil \sqrt{ \frac{T \cdot N_m \cdot \left(T_0 + T_f \right)}{e + e
%\cdot \frac{N_m}{K} + l} } \right \rceil$

%From this result we see that if the number of players leaving or entering is $O(T)$ then we get linear regret, however for all values less than this we get sublinear regret.

\begin{lemma}
Let $e$ be the total number of players entering, $l$  the total number of players leaving, $T_f$ is the expected time for any player to fix on an arm (at most $\exp(2) \cdot N_m $ by lemma \ref{lemma:fixing}), and $\Delta_{min} $ be a lower bound on $\min_{i} \mu_i - \mu_{i+1}$ where $\mu_i$ is the expected reward of the $i^{th}$ best arm, for any $i\leq N_m+1$. 

Then $\forall \delta\in (0,1)$ and $\epsilon < \Delta_{\min}$, w.p. $ \geq 1 - \delta $, the expected regret of the Dynamic MC algorithm played for $ T $ rounds, with parameters: \\
$ T_0 = \left\lceil \max\left( \frac{K}{2} \cdot \ln \left( \frac{2 \cdot K^2}{\frac{\delta}{2 \cdot T}} \right), \frac{16 \cdot K}{ \epsilon^2 } \cdot \ln\left( \frac{4 \cdot K^2}{\frac{\delta}{2 \cdot T}} \right) , \frac{K^2 \cdot \log\left(\frac{4 T}{ \delta}  \right) }{ 0.02 }  \right) \right\rceil $ and $ T_1$ chosen such that $ T_1 > T_0 $, is at most
\[
\frac{T}{T_1} \cdot \left( N_m \cdot \left(T_0 + 2 \cdot T_f \right)\right) + e\cdot 2 \left(T_1 - T_0 \right) + l\left(T_1 - T_0 \right).
\]
\label{lemma:DMC}
\end{lemma}

Note that $N_m$ is not known to the players, however, it is always possible to upper bound it since we are in the setting where the number of players does not exceed the number of arms, i.e., $N_m \leq K$ and thus we can calculate a sufficient time for learning, $T_0$, by replacing $N_m$ by $K$.

The lemma is proven by using lemma~\ref{lemma:mclearn} and lemma~\ref{lemma:unknown} with the confidence parameters set to $ \frac{\delta}{2 \cdot T} $, and taking the union bound over all epochs. This ensures that, with high probability, the players learn the true rankings and estimate the number of players correctly at each epoch. For this reason $ T_0 $ includes a $ \log\left(T \right)  $ factor as stated above. We then separately bound the regret arising from the learning phase and fixing on an arm, as well as regret due to entering and leaving players. The formal proof of this lemma appears in appendix~\ref{app:proofs}.

\section{Comparison to the MEGA Algorithm}\label{sec:mega}
\label{comparison}

As discussed in the introduction, the most relevant existing algorithm for our setting (at least, that we are aware of) is the MEGA algorithm presented in \cite{avner2014concurrent}. In terms of formal guarantees, the algorithm attains $O(T^{2/3})$ in the static setting. 
A full analysis of the algorithm in the dynamic setting is lacking, but it is shown that if a single player leaves at some time point, the system re-stabilizes at an optimal configuration, after essentially $O(T^{2/3})$ rounds. The algorithm is clever, based on well-established techniques, allows players to enter and leave at any round, and compared to our approach, is not based on repeatedly restarting the algorithm, which can be wasteful in practice (an issue we shall return to later on). On the flip side, our algorithms have fewer parameters, attain considerably better performance in the static setting, and can provably cope with the general dynamic setting. In this section, we show that this is not just a matter of analysis, and that the approach taken by the MEGA algorithm indeed has some deficiencies in the dynamic setting. We begin by outlining the MEGA algorithm at a level sufficient to understand our analysis, and then demonstrate how it may perform poorly in some natural dynamic scenarios. 

\subsection{Outline of the algorithm}

The MEGA algorithm uses a well known $\epsilon$-greedy MAB approach, augmented with a collision avoidance mechanism. Initially, players mostly explore arms in order to learn their ranking, and then gradually move to exploiting the best arms, while trying to avoid arms they have collided on. Specifically, each player has an exploration probability which scales like $O\left( \frac{1}{t} \right)$, where $t$ is the current round. The exploration probability also depends on two input parameters, $c$ and $d$, where $d$ is a lower bound on the gap of the $N^{th}$ and $N+1^{th}$ best arms. Each player has a persistence probability, $p_t$, whose initial value, $p_0$, is another input parameter. $p_t$ is increased to $p_{t-1} \cdot \alpha + (1 - \alpha)$ for every round in which the player picks the same arm consecutively, where $\alpha$ is another input parameter. Otherwise, if the player switches arms, $p_t$ is set to $p_0$. In the case of a collision, the colliding players indefinitely flip a coin with their own respective probabilities, $p_t$, for deciding whether to persist on the arm on which they collided. In case a player does not persist after a collision, she marks this arm unavailable until a time point sampled uniformly at random from $\{t, ...,t + t^\beta \} $  where $\beta$ is another input parameter of the algorithm.

Note that both our algorithm and the MEGA algorithm require a lower bound on the gap between the $N^{th}$ best arm and the $N+1^{th}$ best arm.

One issue with the MEGA algorithm approach is that players never entirely stop colliding, even when $T\rightarrow \infty$. At least in the static case, it seems advantageous to fix player's choice after a while, hence avoiding all future collisions and additional regret. The motivation for the MC algorithm is to create a procedure which guarantees that once learning completes, all players will choose one of the $N$ best arms for the rest of the game.

In the dynamic setting, however, the issue is quite the reverse: The $\epsilon$-greedy mechanism, which the MEGA algorithm is based on, is not good at adapting to changing circumstances. In the next subsection, we illustrate two problematic ramifications: One is that players entering late in the game are not able to learn the ranking of the best arm, and the other is that when players leave, the best arms may stay vacant for a long period of time before being sampled by other players. A third issue is that if the reward distributions change over time, a rapidly decreasing exploration probability is problematic. The DMC algorithm can address this as it runs in epochs, hence any mistake in one epoch is undone in the next.

%The MEGA algorithm also relies on a reasonable choice of parameters, but does not provide a clear method for how to choose these parameters other than optimizing, experimentally, over a large set of possible choices. The MC (DMC) algorithm relies on only one (two) parameter(s) and a clear method for calculating theoretically optimal values is provided.

%The MEGA algorithm is based on a combination of the classic $\epsilon-$greedy algorithm and the ALOHA collision avoidance protocol.

\subsection{Problematic Scenarios for the MEGA algorithm}
Below, we study the realistic situation where players both enter and leave, and demonstrate that the regret of the MEGA algorithm can be substantially worse than our regret guarantees (both in terms of regret guarantees as well as in terms of actual regret obtained), sometimes even linear in $T$. For the proofs of the theorems presented in this section we refer the reader to appendix~\ref{app:proofs}.

The first scenario we wish to discuss is the simple setting of two players and two arms, where the second player enters at some round in the game and the first player then leaves at some later round. We will describe what will happen, intuitively, if the players are following the MEGA algorithm, with a formal theorem presented below. In the scenario we described, the first player will learn a correct ranking of the two arms with high probability, and will proceed to exploit the highest ranked arm, thus making his persistence probability very high and exploration probability very low. If a second player enters late in the game, then any attempt to sample the highest ranked arm will cause a collision in which the first player will stay, and the second player will fail to sample the arm, since the first players' persistence probability is so high and the new player's persistence is set to a lower $p_0$. This means that the second player will not be able to learn the true ranking of the two arms. Thus, if the first player leaves after a period of time, such that the second player is not likely to explore, then the second player will exploit the second ranked arm, causing linear regret. This scenario can be extended to multiple players and arms, by adding players one by one in time intervals that ensure that players who entered late will not succeed in learning the true ranking, due to the collision avoidance mechanism. 

The formal result regarding this scenario is the following:
\begin{theorem}
Consider a multi-player MAB setting as described above, where the second player enters at round $\left\lceil \frac{T}{2} \right\rceil$, and the first player leaves at round $\left\lceil \frac{T}{2} + f \cdot T \right\rceil$) for some parameter $f$. Then for all choices of the MEGA algorithm parameters $c,d,\beta,p_0$, if $ \alpha $ (the parameter that controls the collision avoidance mechanism) is chosen such that $ \alpha \leq  1-\frac{4\log(4fT)}{T} $, and $f$ is chosen such that $ \frac{\frac{c \cdot K^2 }{d^2 \cdot  \left( K-1 \right) }}{T} \leq f \leq \frac{d^2 \cdot  \left( K-1 \right) }{8 \cdot c \cdot K^2}$, then:
\begin{itemize}
\item The expected regret of the MEGA algorithm is $ \Omega \left(T \right) $.
\item The conditional expected regret of the DMC algorithm (using $\Theta(\sqrt{T})$ epochs) is $\tilde{O} \left(\sqrt{T} \right) $.
\end{itemize}
\label{thm:discuss2}
\end{theorem}
Notice that for the DMC algorithm we have an upper bound on the expected regret conditioned on the event that all players learn an $\epsilon$-correct ranking, which happens with arbitrarily high probability.
 
In particular, if we choose $f$ to be a constant in the required range, we get a scenario where the regret bounds above hold for any $\alpha \leq 1-O(log(T)/T) $ (and any possible values of the other parameters of the MEGA algorithm). We note that when $\alpha$ is larger than $1-O(log(T)/T)$, the persistence probability $p$ will hardly deviate from $p_0$, which makes the persistence mechanism non-functional and can easily lead to large regret, even in the static setting. 

We now turn to discuss a second reasonable scenario, in which players alternate between entering and exiting, at intervals of $T^\lambda$ rounds. We will show in this scenario that however $\lambda$ is chosen, the regret bound of the MEGA algorithm (as given in \cite{avner2014concurrent}, using recommended parameter values, and even just counting regret due to players leaving) is worse than the regret bound of the DMC algorithm (which incorporate regret due to both players leaving and players entering). Note that unlike Theorem~\ref{thm:discuss2}, here we compare the available regret upper bounds, rather than proving a regret lower bound.

The setting is defined as follows: one player exits (or enters, alternating) every $ T^\lambda $ rounds, for some $\lambda < 1 $. In the worst case, the player who left was occupying the highest ranked arm. In the analysis
of \cite{avner2014concurrent}, players following the MEGA algorithm might take up to $ t^\beta $ rounds before being able to access this arm, due to the collision avoidance mechanism (where $ t $ is the round at which the player
exited, and $\beta $ is a parameter of the algorithm, whose recommended value based on the static setting analysis is $2/3$). For players following the DMC algorithm a player leaving can affect the regret of the current epoch only.
Intuitively, if we set the epoch length compatible to the rate of exiting players, we can achieve a better regret bound than what is indicated by the MEGA algorithm analysis. Formally, we have the following:

%We remind the reader that the MEGA algorithm\footnote{For a detailed
%description of the MEGA algorithm and an explanation of the parameters
%described in the following analysis, please refer to
%\cite{avner2014concurrent}} uses $ p $, the persistence probability, which
%increases while a player remains on some arm. $ p $ determines with what
%probability this player will stay on that arm in the event of a collision. We
%would also like to remind the reader that the probability of exploration goes
%down inversely proportional to the number of rounds that have been played
%thus far.

\begin{figure}[t]
\centering
\includegraphics[scale=0.6]{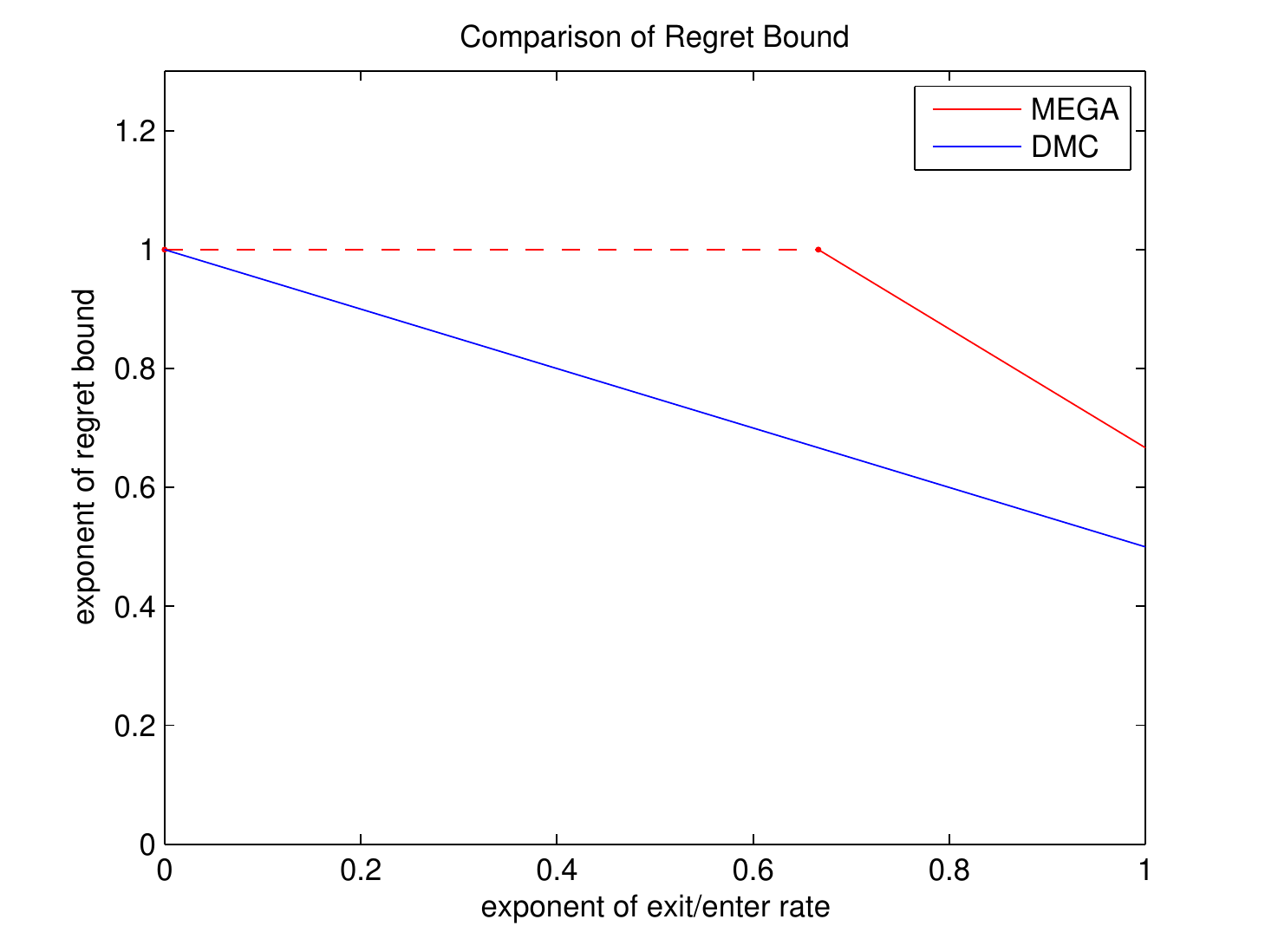}
\caption{The red line represents the exponent $\min\{1,1-(\lambda-\beta)\}$ in the regret upper bound for the MEGA algorithm, as a function of $\lambda$, where we use $\beta=\frac{2}{3}$, the value recommended in \cite{avner2014concurrent}. Note that the dashed red line is a region not covered in the MEGA algorithm analysis, but the regret is trivially at most $T$. The blue line represents the regret exponent $1-\frac{\lambda}{2}$ of the DMC algorithm, as a function of $\lambda$.}
\label{compare_fig}
\end{figure}

%The formal statements of our conclusions regarding this scenario are as follows:
\begin{theorem}
In the multi-player MAB setting with one player leaving every $(2 \cdot r) \cdot
\lceil T^\lambda \rceil$ rounds, and one player entering every $(2 \cdot r + 1)
\cdot \lceil T^\lambda\rceil $ rounds, for $r = 0,1,2,...$, and  $\lambda > \beta$,
we have that
\begin{enumerate}
\item The regret upper bound of the MEGA algorithm is at least
    $O \left( T^{1 - \left( \lambda - \beta \right)} \right)$
\item The expected regret upper bound of the DMC algorithm is
    $\tilde{O} \left( T^{1- \frac{\lambda}{2}} \right)$
\end{enumerate}
\label{thm:discuss1}
\end{theorem}
As in the previous theorem, the expected regret of the DMC algorithm is conditioned on the event that all players learn an $\epsilon$-correct ranking, which happens with arbitrarily high probability.
Also, the assumption $\lambda > \beta$ is required in order to apply the existing MEGA analysis. The theorem is illustrated graphically in Figure~\ref{compare_fig}, and shows how the exponent in the regret bound is uniformly superior for our algorithm, when we pick the recommended value $\beta=\frac{2}{3}$. Note that if $\beta$ is chosen differently then the regret bound of \cite{avner2014concurrent} will increase, even in the static setting.

%Note that the
%DMC upper bound is better than the MEGA algorithm lower bound when $\lambda <
%2 \beta$. This will lead to an upper bound of the DMC regret which is lower
%than the lower bound of the MEGA algorithm. In the static setting the optimal
%value of $\beta$, as in the analysis in \cite{avner2014concurrent}, is
%$\frac{2}{3}$. For this value our bound is always lower than the bound on MEGA. For the dynamic setting we do not have analysis of the MEGA algorithm, thus it is possible that for $\lambda > 2 \beta$ DMC regret will still be lower.

\section{Experiments}\label{sec:experiments}
For our experiments, we implemented the DMC algorithm for the dynamic case and the MC algorithm for the static case. For comparison, we implemented the MEGA algorithm of \cite{avner2014concurrent}, which is the current state-of-the-art for our problem setting.

For each experimental setup and algorithm, we repeated the experiment 20 times, and plotted the average and standard deviation of the resulting regret (the standard deviation is shown with a shaded region encompassing the average regret). In scenarios that are dynamic we mark the time that a player enters or leaves with a dashed line. In most figures, we plot the average per-round regret, as a function of the number of rounds so far.

For the parameters of the MEGA algorithm, we used the empirical values suggested in \cite{avner2014concurrent} (rather than the theoretical values which are overly conservative). The only exception is the gap between the mean rewards of the $N^{th}$ and $N+1^{th}$ best arms, which was taken as the actual gap rather than a rough lower bound. Note that this only gives the MEGA algorithm more power. Moreover, in all experiments, the gap is at least 0.05, which is the heuristic value suggested to be used as the lower bound in \cite{avner2014concurrent}. For the dynamic scenarios, where the players and the number of players change, we use the minimum gap between the $N^{th}$ and $N+1^{th}$ best arms over all rounds. For example, if at the beginning there are 2 players and the gap between the second and third arm is 0.3, but by the end there are 4 players and the gap between the fourth and fifth arm is 0.01, then we use 0.01 as the value of this gap.

For the MC and DMC algorithm, we set $T_0$ to be $3000$ in all experiments. For the DMC parameter, $T_1$, we use either the theoretically optimal value presented in this work or that value scaled by a small constant (see details below for the specific value in each experiment).

In the DMC algorithm, a potential source of waste is that newly entering players can accumulate linear regret until the next epoch begins. Therefore, in the DMC experiments, we added the following heuristic: When a player enters during the middle of the epoch, she chooses an arm with probability proportional to the empirical mean of its rewards (as observed by her so far, initially set to 1), multiplied by the empirical probability of not colliding on that arm (initially set to 1). After the epoch is over, she chooses arms by following the DMC algorithm. Intuitively this would quickly stop a large amount of collisions with players who are already fixed, and would encourage more players to exploit rather than only explore. This happens because any arm that has a player `fixed' on it would always give newly entered players a zero empirical probability for colliding. 
%For any arm that does not have a player fixed on it, the probability of colliding is not very meaningful. If there are multiple newly entered players then the probability of colliding should go up for arms that have higher expected reward.

We begin with a simple scenario corresponding to the static setting. There is an initial set of 6 players, which remains fixed throughout the game, and 10 arms. The mean rewards of the arms are chosen uniformly at random in $[0,1]$ (with a gap of at least $0.05$ between the $N^{th}$ and $N+1^{th}$ arm). At every round of the game the rewards of each arm are chosen to be $1$ with probability equal to the mean reward, and zero otherwise.

%\begin{figure}[ht]
%\hspace{-20mm}
%\begin{subfigure}[h]{0.3\textwidth}
%\centering
%\includegraphics[scale=0.7]{static_50k_rand.png}

%\caption{average regret}
%\label{static_rand}
%\end{subfigure}
%\hspace{45mm}
%\begin{subfigure}[h]{0.3\textwidth}
%\centering
%\includegraphics[scale=0.7]{static_50k_rand_accum_reg.png}
%\caption{total accumulated regret}
%\label{accum_regret}
%\end{subfigure}
%\caption{In this scenario there are 6 players, 10 arms, and the rewards are chosen uniformly at random. The figure on the left shows the average regret after 50,000 iterations and the figure on the right shows the accumulated regret after 50,000 iterations.}

%\end{figure}

\begin{figure}
\hspace{-8mm}
%    \centering
    \begin{subfigure}{0.5\textwidth}
        \centering
        \includegraphics[width=8cm]{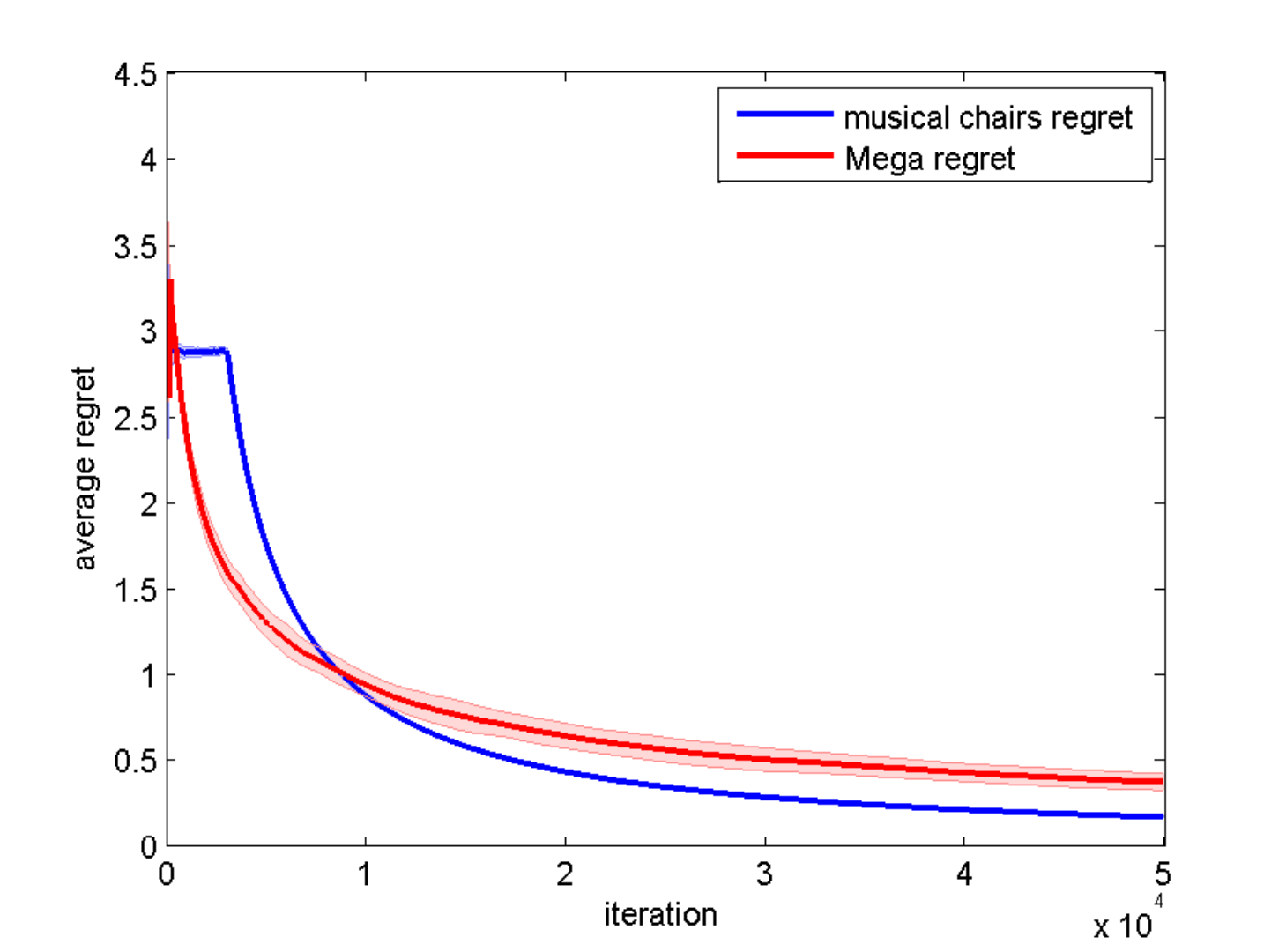}
        \caption{average regret}
        \label{static_rand}
    \end{subfigure}%   %% This % is needed when you use 0.5\textwidth
\hspace{10mm}
    %  Don't leave the blank line
    \begin{subfigure}{0.5\textwidth}%[H]   %% Don't put this here
        \centering
        \includegraphics[width=8cm]{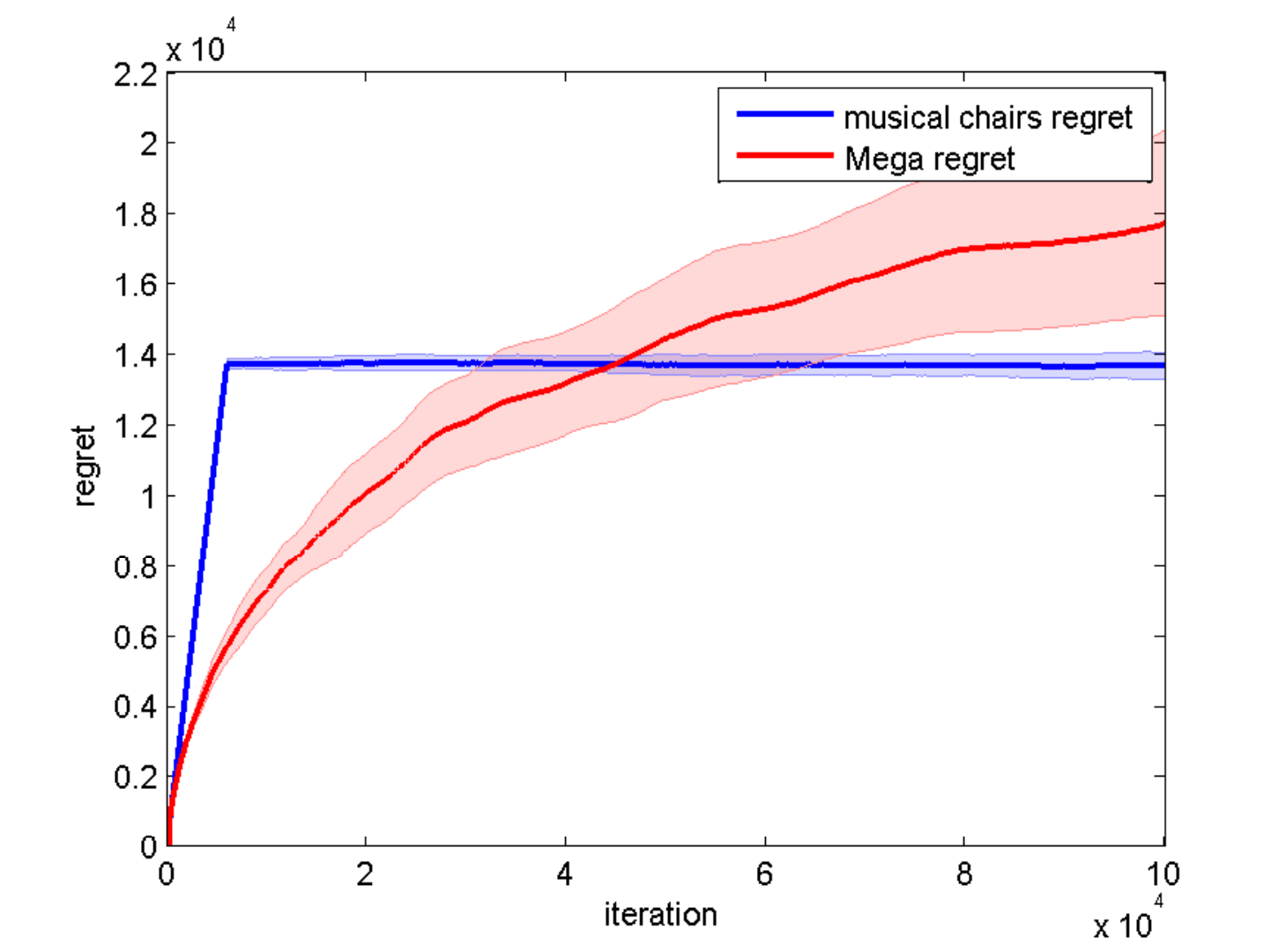}
        \caption{total accumulated regret}
        \label{accum_regret}
    \end{subfigure}
 \caption{The figure on the left shows the average regret after 50,000 iterations and the figure on the right shows the accumulated regret after 50,000 iterations.}
\end{figure}

In this scenario we can see the short time period where players running the MC algorithm are learning (and the average regret is constant), and then, with high probability, they all know which are the best $N$ arms and never make any more mistakes or collisions. The average regret is shown in figure~\ref{static_rand}. The added regret at every round after learning is zero, while in the MEGA algorithm, even though the exploration probability goes down with time, it is never zero. Also, in the MEGA algorithm every time a player has the best arm become 'available' that player will try to exploit it, probably colliding with other players who also want to exploit that arm. Therefore, in the MEGA algorithm there will always be collisions, even though they happen less frequently with time. This is further exemplified in figure~\ref{accum_regret} where we can see that after the learning stage there is no further accumulated regret for the MC algorithm while the MEGA algorithm never stops accumulating regret.

\begin{figure}[t]
	\hspace{-8mm}
	%    \centering
	\begin{subfigure}{0.5\textwidth}
		\centering
		\includegraphics[width=8cm]{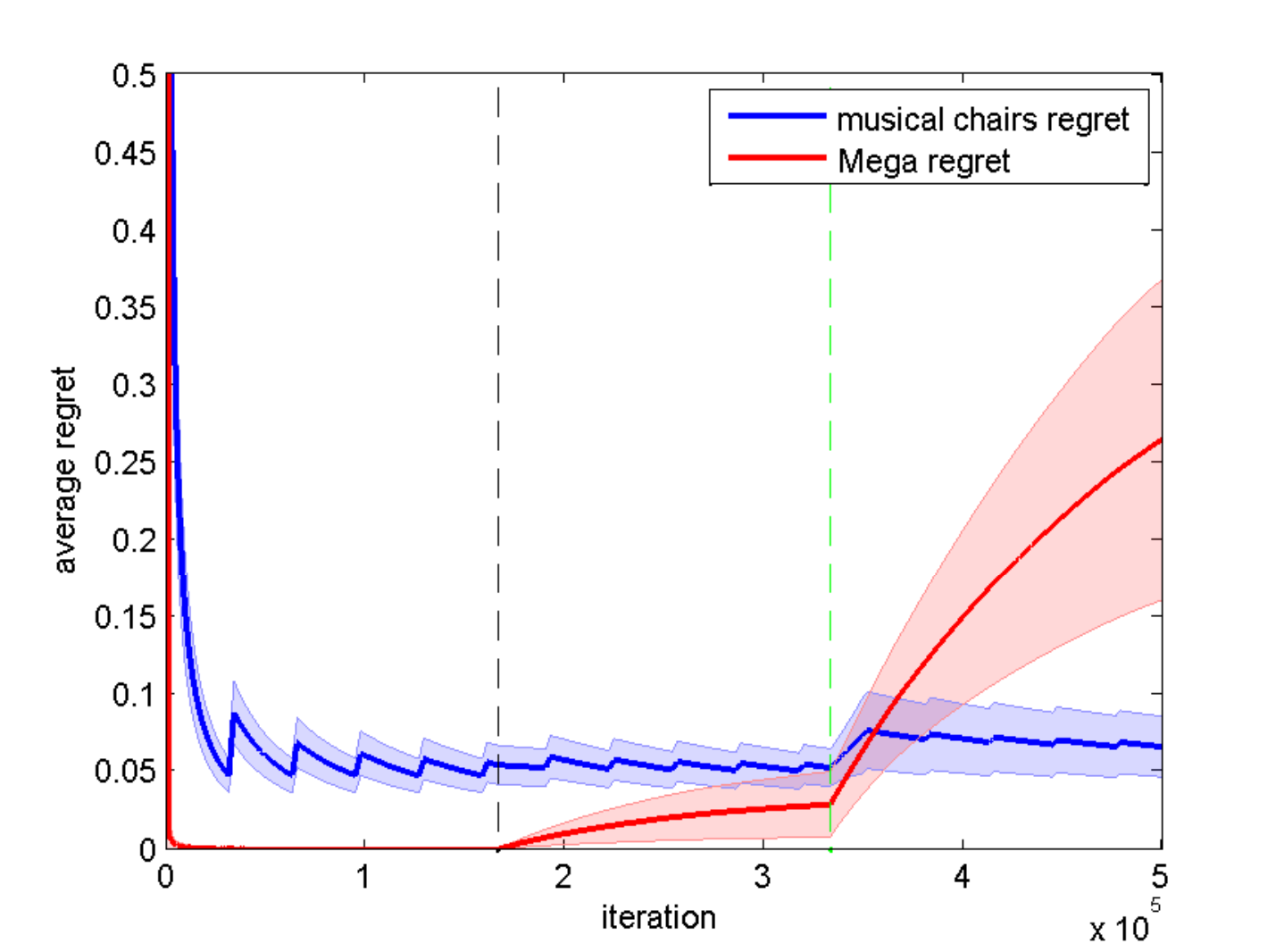}
		\caption{case 1}
		\label{case_1}
	\end{subfigure}%   %% This % is needed when you use 0.5\textwidth
	\hspace{10mm}
	%  Don't leave the blank line
	\begin{subfigure}{0.5\textwidth}%[H]   %% Don't put this here
		\centering
		\includegraphics[width=8cm]{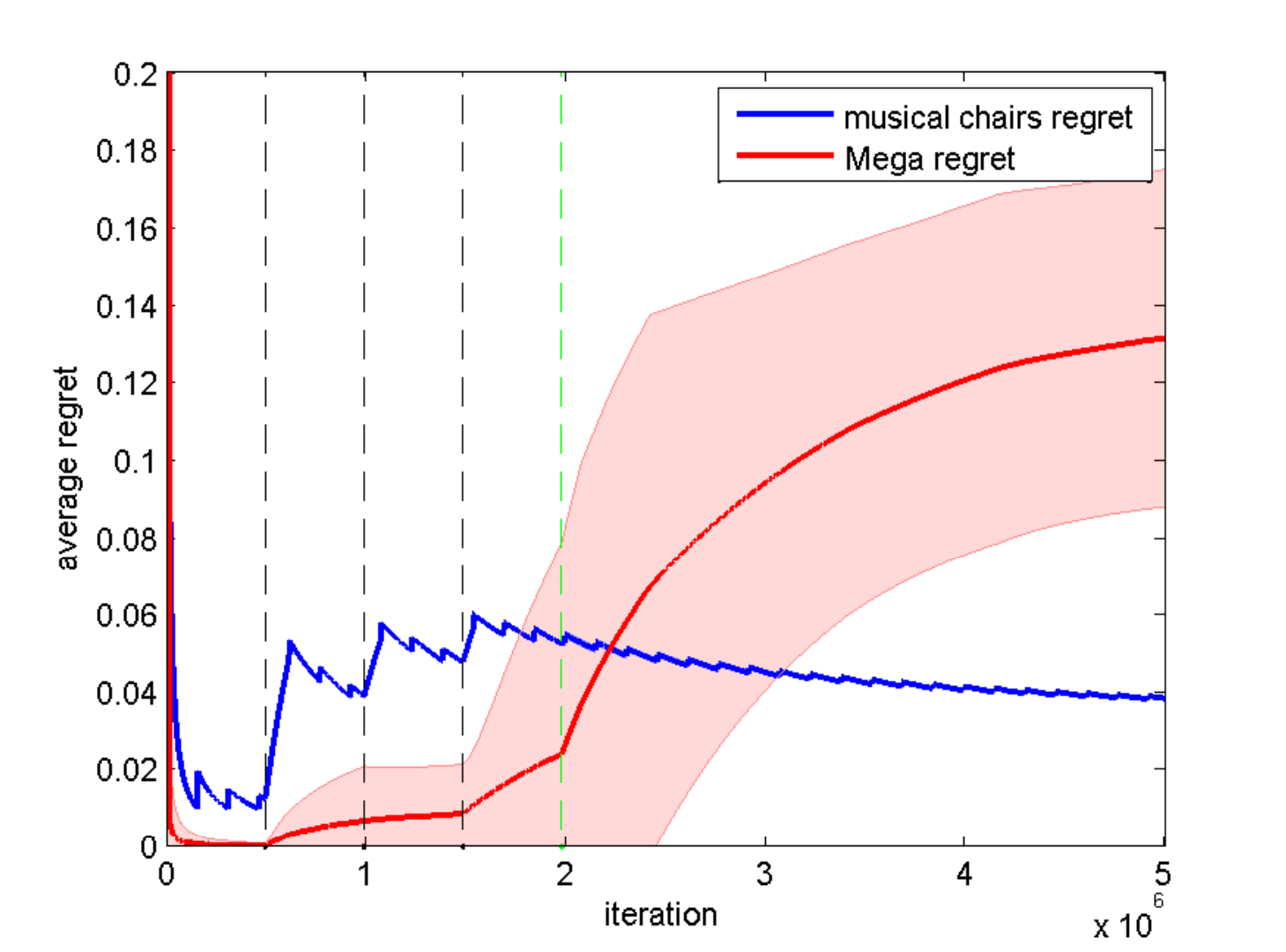}
		\caption{generalized case 1}
		\label{general_case_1}
	\end{subfigure}
	\caption{The black dotted lines represents players entering and the green dotted line shows when the first player exits.}
\end{figure}

\begin{figure}[t]
	\hspace{-8mm}
	%    \centering
	\begin{subfigure}{0.5\textwidth}
		\centering
		\includegraphics[width=8cm]{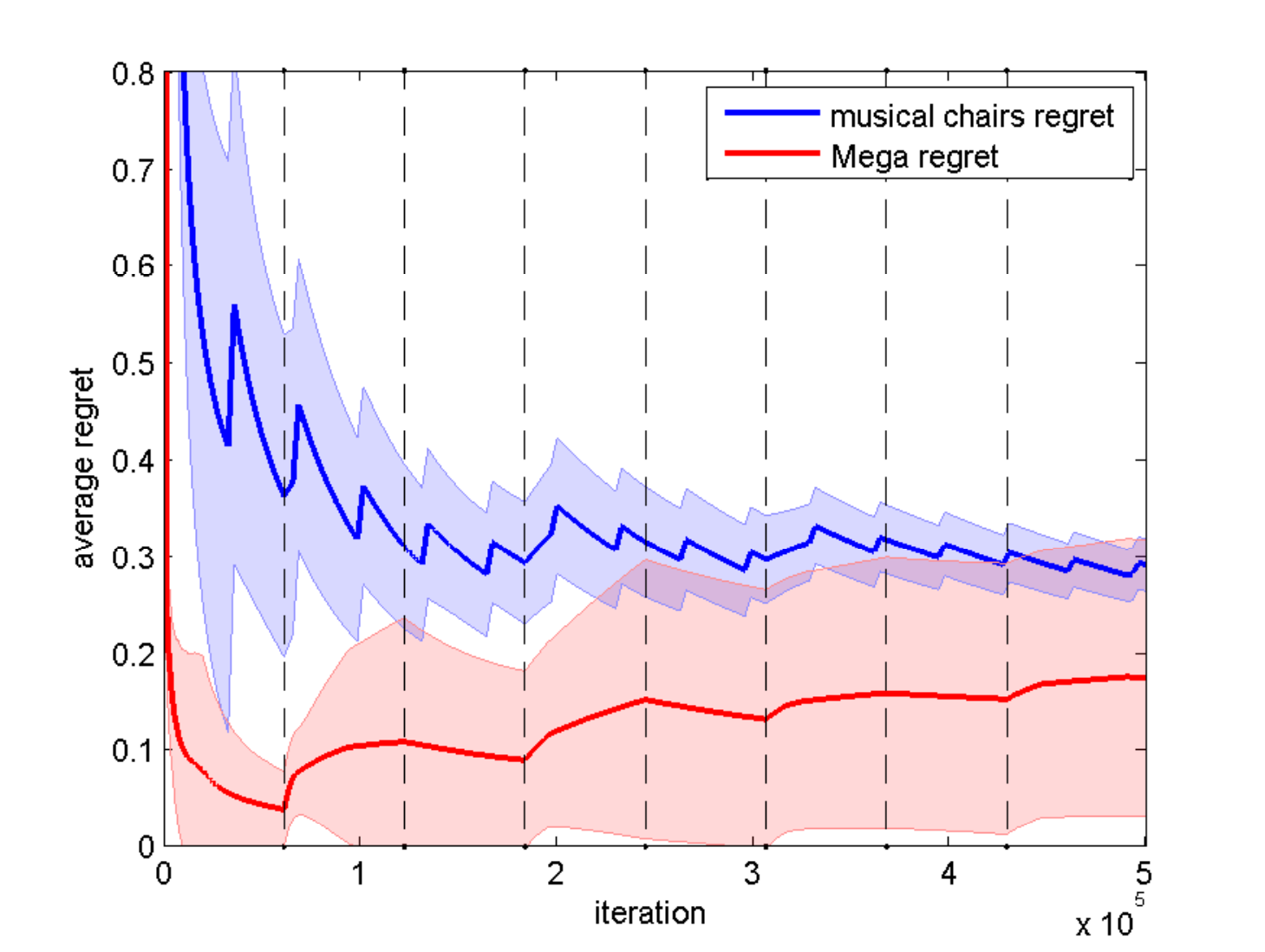}
		\caption{case 2}
		\label{case_2}
	\end{subfigure}%   %% This % is needed when you use 0.5\textwidth
	\hspace{10mm}
	%  Don't leave the blank line
	\begin{subfigure}{0.5\textwidth}%[H]   %% Don't put this here
		\centering
		\includegraphics[width=8cm]{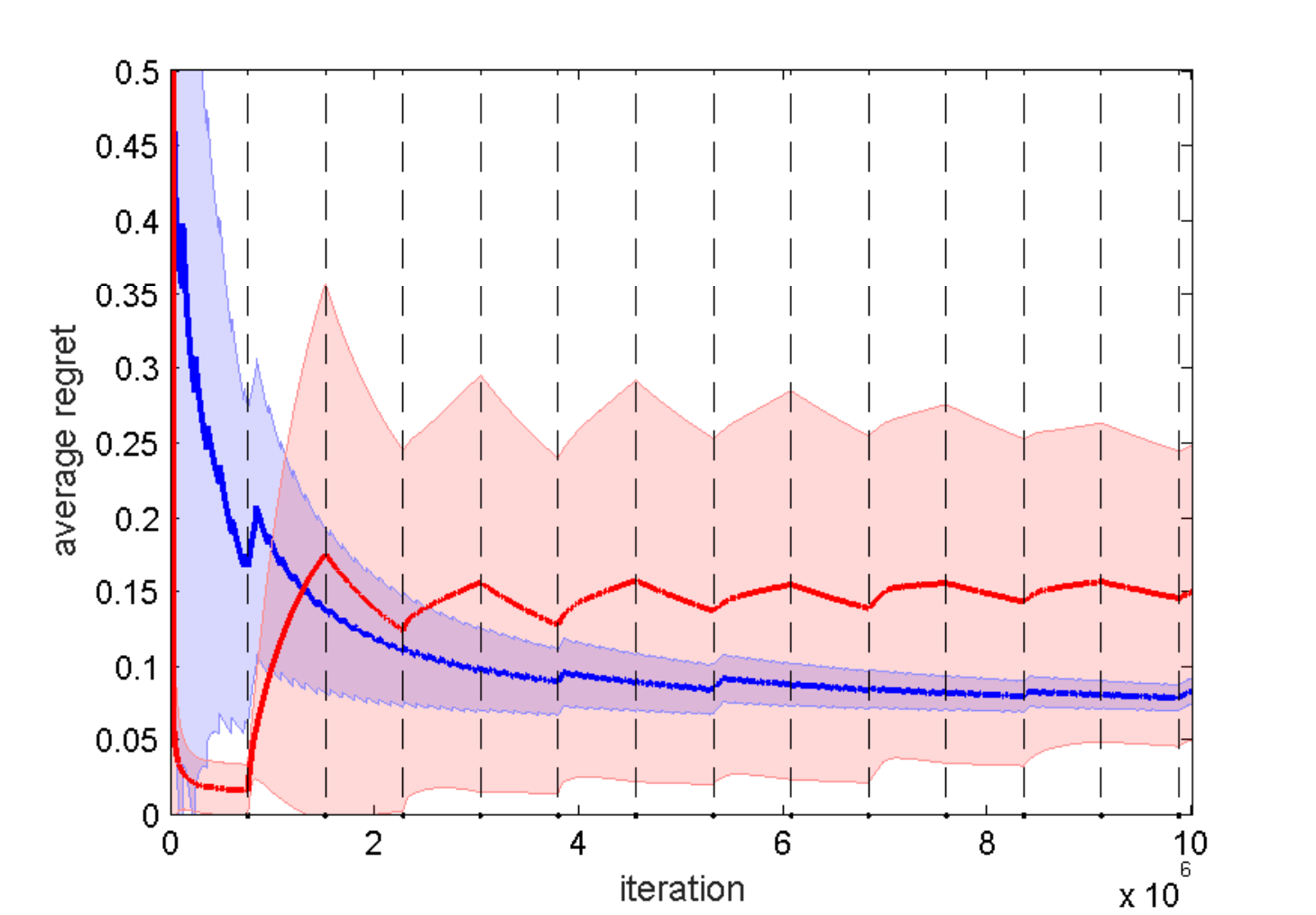}
		\caption{case 2 with 10 million iterations}
		\label{10_mil}
	\end{subfigure}
	\caption{Players alternate between entering and leaving (black dotted lines)}
\end{figure}

Another scenario we simulate is that of section~\ref{comparison} in theorem~\ref{thm:discuss2}. The game starts with one player. At round $\lceil \frac{T}{3} \rceil$, a second player enters and after another $\lceil \frac{T}{3} \rceil$ rounds the first player leaves. The results can be seen in figure~\ref{case_1}. In figure~\ref{general_case_1} we show a generalization of this scenario for multiple players where the game starts with one player and every $T^{0.84}$ rounds another player enters, until the number of players reaches 4. Then after another $T^{0.84}$ rounds, the first player leaves. For both of these scenarios the rewards are chosen deterministically. For figure~\ref{case_1} there is a lower bound on the gap of 0.8 and for figure~\ref{general_case_1} there is a gap of 0.7 between the expected reward of the $N^{th}$ and $N+1^{th}$ best arm. In figure~\ref{case_1}, there are 4 arms and $T_1$ was set to $34,757$, and in figure~\ref{general_case_1}, there are 10 arms and $T_1$ was set to $167,845$.
%is set to the theoretical value suggested in theorem~\ref{thm:DMC} times $1.26$, which is $34,757$. In figure~\ref{general_case_1} $T_1$ is set to $9.6$ times the theoretical value which is $167,845$. 

As discussed in section~\ref{sec:mega}, in the scenario of figure~\ref{case_1} a second player who enters late is not able to learn the best arm, because the first player always exploits the best arm and has a very high persistence probability. Therefore, once the first player leaves the game, and allows the best arm to be free for the second player to use, it will take the second player time proportional to the number of rounds she has played to explore this arm. Since her exploration probability will be very low, exploring this arm will take a very long time. The DMC algorithm runs in epochs and therefore a problem of inflexibility, or an inability to change for dynamic settings, does not arise. This phenomenon can be seen in figures \ref{case_1} and \ref{general_case_1}: When the first player leaves (marked by a dotted green line) the average regret of the MEGA algorithm increases dramatically, while for the DMC algorithm the decreasing trend continues. This suggests that the MEGA algorithm may be susceptible to large regret in some natural dynamic scenarios.

%\begin{figure}[ht]
%\hspace{-20mm}
%\begin{subfigure}[h]{0.3\textwidth}
%\centering

%\includegraphics[scale=0.7]{case_1_4_arms.png}

%\caption{case 1}
%\label{case_1}
%\end{subfigure}
%\hspace{45mm}
%\begin{subfigure}[h]{0.3\textwidth}
%\centering
%\includegraphics[scale=0.7]{case_2_5_player_10_arms.png}

%\caption{case 2}
%\label{case_2}
%\end{subfigure}
%\caption{In figure~\ref{case_1} there are 4 arms, one player starting, a second entering at round $\lceil \frac{T}{3} \rceil$, the first leaves after another $\lceil \frac{T}{3} \rceil$ rounds. The black dotted line represents the second player entering and the green dotted line shows when the first player exits. In figure~\ref{case_2} the game starts with 5 players, 10 arms, and there is one player leaving/entering every 69843 rounds marked by black dotted lines.}
%\end{figure}

Another dynamic player scenario we simulate (demonstrating theorem~\ref{thm:discuss1} in section~\ref{sec:mega}) is where the game starts with a set of five players and 10 arms and every $T^{0.84}$ rounds, we alternate between a player leaving and a player entering. The leaving player is chosen at random from the set of current players. Figure~\ref{case_2} shows the outcome for $T=5*10^5$, and \ref{10_mil} shows the outcome for $T=6*10^6$. In these scenarios $T_1$ is chosen to be
$32,482$ in figure~\ref{case_2} and  $119,921$ in figure~\ref{10_mil}.
% $2.19$ (for figure~\ref{case_2}) and $2.47$ (for figure~\ref{10_mil}) times the theoretical value suggested in theorem~\ref{thm:DMC} (which is $32,482$ in figure~\ref{case_2} and $119,921$ in figure~\ref{10_mil}). 
Although our algorithm performs better when $T$ is large enough (confirming the theoretical evidence in theorem \ref{thm:DMC}), we note that this is not the case for smaller values of $T$. We believe that this is due to the epoch-based nature of the DMC algorithm, which can be wasteful when $T$ is moderate. However, when $T$ is sufficiently large (as in figure~\ref{10_mil}), the DMC algorithm outperforms the MEGA algorithm.

\section{Discussion}\label{sec:discussion}

In this work we propose new algorithms for the stochastic multi-player multi-armed bandit problem, with no communication or central control. We provide an analysis for the static setting, showing that the proposed MC algorithm achieves a better upper bound on the regret (as a function of the number of rounds) than the current state of the art. We also provide the DMC algorithm, which is the first (to the best of our knowledge) with formal guarantees which copes with the general dynamic setting. We also study some natural dynamic scenarios, in which the behavior of previous approaches is problematic, sometimes even leading to linear regret.

This work leaves several questions open. For example, as noted earlier, both the DMC algorithm and the earlier MEGA algorithm require knowing a lower bound on the gap between the $N^{th}$ best arm and the $N+1$ best arm, and it would be interesting to remove this assumption while attaining similar guarantees. Another issue with the DMC algorithm is its epoch-based nature, which in practice considerably degrades the regret (especially if the total number of rounds $T$ is not too large). Can we develop algorithms with provable guarantees for the general dynamic setting which are not epoch-based?

%A more minor issue of the DMC algorithm is that new players are not handled well. As currently written, the algorithm suggests that newly entering players will randomly explore until the end of the epoch, which makes the regret fragile and dependent on the epoch length. As detailed in the experiments section, it is possible to augment the DMC algorithm by a simple heuristic, under which new players choose arms in a way reflecting their empirical mean and collision probability, at least till the end of the current epoch. 

More generally, there are several interesting variants of the multi-player MAB setting, that are currently unexplored. For example, it would be quite interesting to develop algorithms for multi-player MAB in the non-stochastic (adversarial) setting, where the rewards are arbitrary. In the adversarial case, one cannot rely on high-reward arms to remain such in the future, and it is not clear at all what algorithmic mechanism can work here. Another interesting direction is to remove the assumption that players faithfully execute a given algorithm: In practice, players may be non-cooperative and greedy, and it would be interesting to devise algorithms which are also incentive-compatible, and study related game-theoretic questions.

\subsubsection*{Acknowledgments}
This research is partially supported by Israel Science
Foundation grant 425/13, and an FP7 Marie Curie CIG grant. We thank Nicol\`{o} Cesa-Bianchi and Yishay Mansour for several discussions which helped initiate this line of work.

\bibliographystyle{plainnat}
\bibliography{biblio}

\newpage

\appendix
\section{Proofs}
\label{app:proofs}

We will use the following standard concentration bounds:

Let $X_1, ..., X_m$ be independent random variables such that $X_i$ always lies in the interval $[0,1]$. Define $\overline{X} = \frac{1}{m} \sum_{i=1}^m X_i$ and $\mu = \mathbb{E}[\overline{X}]$.
\begin{theorem}(Chernoff Bound)
For any $0 < \delta < 1$,
\begin{gather*}
\Pr \left( \sum_{j=1}^m X_j \geq (1-\delta) \cdot \mathbb{E}[\sum_{j=1}^m X_j] \right) \leq \exp\left(-\frac{\mathbb{E}[\sum_{j=1}^m X_j] \delta^2}{2}\right)
\end{gather*}
\end{theorem}

\begin{theorem}(Hoeffding's Inequality)
For any $\delta > 0$,
\begin{gather*}
\Pr \left( |\overline{X} - \mu |\geq \delta \right) \leq 2 \cdot \exp\left(-2 \cdot m \cdot \delta^2\right)
\end{gather*}
\end{theorem}

\subsection{MC algorithm proofs}

We prove the following lemmas from which Theorem~\ref{thm:MC} follows.

\subsubsection{Proof of Lemma~\ref{lemma:mclearn} }
%We begin by restating the lemma:\\
%\textbf{Lemma~\ref{lemma:mclearn}.}
%\textit{
%$ \forall \epsilon > 0 $ and $ 0<\delta <1 $ there exists a $ T_0 = \left\lceil \max \left\{ \frac{K}{2} \cdot \ln \left( \frac{2 \cdot K^2}{\delta} \right), \frac{16 \cdot K}{\epsilon^2} \cdot \ln\left( \frac{4 \cdot K^2}{\delta} \right) \right\} \right\rceil$ s.t. after $ T_0 $ rounds of random exploration all players have an $ \epsilon- $correct ranking of the arms w.p. $ \geq 1 - \delta $
%}
%\begin{proof}
We will first upper bound the probability that not all players have an $ \epsilon- $correct ranking given that they have enough observations of each arm. Then we upper bound the probability that not all players have enough observations of each arm, given that they have played $ T_0 $ rounds. We then take a union bound over both of these bad events to upper bound the probability that not all players have an $ \epsilon- $correct ranking after $ T_0 $ rounds of random exploration.\\
Given $ \delta $ we define $ \delta_1 := \frac{\delta}{2} $ and $ \delta_2 := \frac{\delta }{2} $.\\
Note that if for player $ i $ it is true that $ \forall j \in \{1,...,K\} $ $ |\tilde{\mu}_j - \mu_j | \leq \frac{\epsilon }{2}$ then player $ i $ must have an $ \epsilon- $correct ranking.\\
We now calculate what is the required number of observations, $ C $, of each arm in order to get
\begin{gather*}
\Pr \left( \mbox{player $ i $ doesn't have an $ \epsilon- $correct ranking $ | $ player $ i $ viewed $ \geq C $ observations of each arm} \right) < \frac{\delta_1}{N}.
\end{gather*}
Specifically, we have the following:
\begin{gather*}
\Pr \left( \mbox{player $ i $ does not have an $ \epsilon- $correct ranking $ | $ player $ i $ viewed $ \geq C $ observations of each arm} \right)\\ \leq \Pr \left( \exists j \mbox{ s.t. } |\tilde{\mu}_j - \mu_j | > \frac{\epsilon }{2} \mbox{ $ | $  player $ i $ viewed $ \geq C $ observations of each arm} \right)\\ \underset{\mbox{union bound}}{\underbrace{\leq}} \sum_{j=1}^K  \Pr \left( |\tilde{\mu}_j - \mu_j | > \frac{\epsilon }{2} \mbox{ $ | $  player $ i $ viewed $ \geq C $ observations of each arm} \right)\\
= \sum_{j=1}^K \sum_{n=C}^{\infty} \Pr \left(|\tilde{\mu}_j - \mu_j|> \frac{\epsilon}{2} | \mbox{ \# of views }  = n \right)\Pr\left(\mbox{ viewed } n | n>=C \right)
\end{gather*}
Using Hoeffding's inequality, this is at most
\begin{gather*}
\underset{\mbox{Hoeffding's Inequality}}{\underbrace{\leq}}\sum_{j=1}^K \sum_{n=C}^{\infty} 2 \cdot \exp\left(\left(\frac{-n \cdot \epsilon^2}{2}\right)\right) \Pr \left(\mbox{ viewed } n|n>=C \right)
\\
\leq \sum_{j=1}^K 2 \cdot \exp\left( \left( \frac{-C \cdot \epsilon^2}{2} \right)\right) \sum_{n=C}^{\infty}\Pr \left(\mbox{ viewed } n|n>=C \right)
\\
= K \cdot 2 \cdot \exp\left(\left(\frac{-C \cdot \epsilon^2}{2} \right)\right)
\end{gather*}
Notice that we can apply Hoeffding's inequality here since each observation of the reward of an arm is sampled independent of the number of times we view it. This is true since every player is sampling uniformly at random at every round of learning (and independent of all previous rounds).\\
In order for this to be $ < \frac{\delta_1}{N} $ we need:
\begin{gather*}
2 \cdot K \cdot \exp\left(-C \cdot \frac{\epsilon^2 }{2} \right) < \frac{\delta_1}{N}\\
\implies C > \ln \left( \frac{2 \cdot K \cdot N}{\delta_1} \right) \cdot \frac{2}{\epsilon^2}
\end{gather*}
Now we show that if all players have at least $ C > \ln \left( \frac{2 \cdot K \cdot N}{\delta_1} \right) \cdot \frac{2}{\epsilon^2} $ observations of each arm then w.p. $ \geq 1 - \delta_1 $ all players have an $ \epsilon- $correct ranking:\\
We start by defining the following events:\\
\begin{itemize}
\item $A$ will denote the event that all players have an $\epsilon-$correct ranking ($\overline{A} $ will denote A complement)
\item $A_i$ will denote the event that player $i$ has an $\epsilon-$correct ranking
\item $B$ will denote the event that all players have observed each arm at least $C$ times ($\overline{B} $ will denote B complement)
\item $B_i$ will denote the event that player $i$ has observed each arm at least $C$ times
\end{itemize}

\begin{gather*}
\Pr \left( A | B \right) \\
\geq 1 - \Pr \left( \bigvee_i \overline{A}_i | B_i \right) \\
\underset{\mbox{union bound}}{\underbrace{\geq}} 1 - \sum_{i=1}^N \Pr \left( \overline{A}_i | B_i \right) \\
\geq 1 - N \cdot \frac{\delta_1}{N} = 1 - \delta_1
\end{gather*}
Now we show that there exists a $ T_0 $ large enough so that all players have $ > C $ observations of each arm w.p. $ \geq 1 - \delta_2 $.\\
We define $ A_{i,j} (t) = I \left\{ \mbox{player $ i $ observed arm $ j $ at round $ t $} \right\} $.\\
Note that for any round $t $ and any $ i,j $ we have that $ \Pr \left(A_{i,j} (t)=1 \right) = \frac{1}{K} \cdot \left(1 - \frac{1}{K} \right)^{N-1} $ $ \implies \mathbb{E} \left[ A_{i,j} (t) \right] =  \frac{1}{K} \cdot \left( 1 - \frac{1}{K} \right)^{N-1}$.\\
So for any $ i,j $ we have that
\begin{gather*}
\Pr \left( \mbox{player $ i $ has $ \leq \frac{1}{2} \cdot T_0 \cdot \mathbb{E} \left[ A_{i,j} (t) \right] \mbox{ observations}$} \right) \\ = \Pr \left( \sum_{t=1}^{T_0} A_{i,j} \left(t \right) \leq \frac{1}{2} \cdot T_0 \cdot \mathbb{E} \left[ A_{i,j} (t) \right] \right) \\
\underset{\mbox{Chernoff bound}}{\underbrace{\leq}} e^{\frac{-\frac{1}{4} \cdot T_0 \cdot \mathbb{E} \left[ A_{i,j} (t) \right] }{2} }
\end{gather*}
Note that we can apply Chernoff bound here since for any i,j, $ A_{i,j} $ are i.i.d across t, since all players are employing random sampling at every round of learning.\\
Using a union bound we get that:
\begin{gather*}
\Pr \left( \exists i,j s.t. \sum_{t=1}^{T_0} A_{i,j} \left(t \right) \leq \frac{1}{2} \cdot T_0 \cdot \mathbb{E} \left[ A_{i,j} (t) \right] \right) \\
\leq N \cdot K \cdot \exp\left(\frac{-\frac{1}{4} \cdot T_0 \cdot \mathbb{E} \left[ A_{i,j} (t) \right] }{2} \right)
\end{gather*}
In order for this probability to be upper bounded by $ \delta_2 $ we need:
\begin{gather*}
N \cdot K \cdot \exp\left(\frac{-\frac{1}{4} \cdot T_0 \cdot \mathbb{E} \left[ A_{i,j} (t) \right] }{2} \right) < \delta_2\\
\implies T_0 > \frac{1}{8 \cdot \mathbb{E} \left[ A_{i,j} (t) \right]} \cdot \ln \left( \frac{N \cdot K}{\delta_2} \right)
\end{gather*}
We have shown that if $ T_0 > \frac{1}{8 \cdot \mathbb{E} \left[ A_{i,j} (t) \right]} \cdot \ln \left( \frac{N \cdot K}{\delta_2} \right) $ then w.p. $ \geq 1 - \delta_2 $ we have $ \forall i,j $ the number of observations player $ i $ has of arm $ j $, $ \sum_{t=1}^{T_0} A_{i,j} \left( t \right)$, $ > \frac{1}{2} \cdot T_0 \cdot \mathbb{E} \left[ A_{i,j} (t) \right] $.\\
We also need the total number of observations each player has of each arm to be at least $ C $,
\\
i.e.
\begin{gather*}
\sum_{t=1}^{T_0} A_{i,j} \left( t \right) > \frac{1}{2} \cdot T_0 \cdot \mathbb{E} \left[ A_{i,j} (t) \right] \geq C > \ln\left( \frac{2 \cdot K \cdot N}{\delta_1} \right) \cdot \frac{2}{\epsilon^2} \\
\implies T_0 \geq 2 \cdot \frac{1}{\mathbb{E} \left[ A_{i,j} (t) \right]} \cdot \ln\left( \frac{2 \cdot K \cdot N}{\delta_1} \right) \cdot \frac{2}{\epsilon^2}
\end{gather*}
So we have two constraints on $ T_0 $, thus we take:\\
$ T_0 = \left\lceil \max \left\{ \frac{1}{8 \cdot \mathbb{E} \left[ A_{i,j} (t) \right]} \cdot \ln \left( \frac{N \cdot K}{\delta_2} \right), 2 \cdot \frac{1}{\mathbb{E} \left[ A_{i,j} (t) \right]} \cdot \ln\left( \frac{2 \cdot K \cdot N}{\delta_1} \right) \cdot \frac{2}{\epsilon^2} \right\} \right\rceil$ .\\
We remind the reader that $ \mathbb{E} \left[ A_{i,j} (t) \right] =  \frac{1}{K} \cdot \left( 1 - \frac{1}{K} \right)^{N-1} \geq \frac{1}{K} \cdot \frac{1}{4} $ for all $K > 1$.\\
So we take $T_0 = \left\lceil \max \left\{ \frac{K}{2} \cdot \ln \left( \frac{N \cdot K}{\delta_2} \right), \frac{16 \cdot K}{\epsilon^2} \cdot \ln\left( \frac{2 \cdot N \cdot K}{\delta_1} \right) \right\} \right\rceil$ and the result holds.
Using the events, $A$ and $B$ as defined above, we get that
\begin{gather*}
\Pr \left( A \right) = 1 - \Pr\left( \overline{A}\right)\\
= 1 - \left( \Pr \left( \overline{A}|B \right) \cdot \Pr \left(B \right) + \Pr \left( \overline{A}|\overline{B} \right) \cdot \Pr \left(\overline{B} \right) \right)\\
\geq 1 - \left(  \Pr \left( \overline{A}|B \right)  + \Pr \left(\overline{B} \right) \right)\\
\geq 1 - \left( \delta_1 + \delta_2 \right) \geq 1 - \delta
\end{gather*}

Notice that letting $ T_0 = \left\lceil \max \left\{ \frac{1}{8 \cdot \mathbb{E} \left[ A_{i,j} (t) \right]} \cdot \ln \left( \frac{N \cdot K}{\delta_2} \right), 2 \cdot \frac{1}{\mathbb{E} \left[ A_{i,j} (t) \right]} \cdot \ln\left( \frac{2 \cdot K \cdot N}{\delta_1} \right) \cdot \frac{2}{\epsilon^2} \right\} \right\rceil $ is only possible if one knows know $N$. If $N$ is unknown, then one can increase $T_0$ and set it to 
\[
T_0 = \left\lceil \max \left\{ \frac{1}{8 \cdot \mathbb{E} \left[ A_{i,j} (t) \right]} \cdot \ln \left( \frac{ K^2}{\delta_2} \right), 2 \cdot \frac{1}{\mathbb{E} \left[ A_{i,j} (t) \right]} \cdot \ln\left( \frac{2 \cdot K^2 }{\delta_1} \right) \cdot \frac{2}{\epsilon^2} \right\} \right\rceil,
\] 
and the lemma would still hold since we assume that $N < K$.

%\end{proof}

\subsubsection{Proof of Lemma~\ref{lemma:unknown} }
%We begin by restating the lemma:\\
%\textbf{Lemma~\ref{lemma:unknown}.} \textit{Let $ \delta_2 \in [0,1] $. For $ \epsilon_1 = \frac{0.1}{K}$ if the number of rounds used to estimate $N$ is at least $ T_e = \left\lceil \frac{\log\left(\frac{2}{ \delta_2}  \right) }{2{\epsilon_1}^2 } \right\rceil $, then w.p. $ \geq 1 - \delta_2 $ we have that $ N^{\ast} = N $.}
%\begin{proof}

Fix some player $i$, and let $C_t$ be the number of collisions observed by the player until time $ t $. Also, let $ p $ be the true probability of a collision, when $N$ players are choosing arms uniformly at random among $K$ arms. The probability of a player not experiencing a collision is
\begin{gather*}
\Pr  \left( \mbox{no collision} \right)  = \sum_{j=1}^K \Pr \left( \mbox{choose arm j} \right) \cdot \Pr \left( \mbox{no other player chooses arm j} \right) \\
= \sum_{j=1}^K \frac{1}{K} \cdot  \left( 1 - \frac{1}{K} \right) ^{N-1}\\
= \frac{1}{K}\cdot K \cdot  \left( 1 - \frac{1}{K} \right) ^{N-1} =  \left( 1 - \frac{1}{K} \right) ^{N-1}
\end{gather*}
Thus the probability of a collision at any round of learning is : $ p = 1 -  \left( 1 - \frac{1}{K} \right) ^{N-1} $. Note that $ p < 1 $ for any $ N, K > 0 $. Inverting this equation, we get
\[
N = \frac{\log(1-p)}{\log\left(1-\frac{1}{K}\right)}+1.
\]
Therefore, if we let $ \hat{p}_t := \frac{C_t}{t}$ be the empirical estimate of the collision probability after $t$ rounds, it is natural to take the estimator defined as
\[
N^{\ast} ~=~ \text{round}\left(\frac{\log \left(1- \hat{p}_t \right) }{\log \left( 1-\frac{1}{K} \right)  } + 1\right)~=~\text{round}\left(\frac{\log \left( \frac{t-C_t}{t} \right) }{\log \left( 1-\frac{1}{K} \right)  } + 1\right)
\]
Our goal will be to show that when $t$ is sufficiently large, $N^\ast=N$ with arbitrarily high probability. Specifically, we will upper bound the probability of the estimator $\frac{\log \left( \frac{t-C_t}{t} \right) }{\log \left( 1-\frac{1}{K} \right)  } + 1$ being far from the true value $N$ (which also includes the unlikely case $C_t=t$, in which case the estimator is infinite).

%Note that the probability that $ C_t = t $ is $  \left( 1 -  \left( 1 - \frac{1}{K} \right) ^{N-1} \right) ^t $  (since it is the probability of a having a collision, for t rounds of learning).

%Specifically, let $ \gamma \geq 0 $, $ \delta_2 \in [0,1] $. We show that $ \exists \epsilon_1 \left( N, K, \gamma \right) $ s.t. if $ T_e \geq \frac{\log \left( \frac{\delta_2}{2}  \right) }{-2{\epsilon_1}^2 } $, where $T_e$ is the number of rounds, then w.p. $ \geq 1 - \delta_2 $ we have that $ |N^{\ast} - N | \leq \gamma $.

Recalling that $N=\frac{\log(1-p)}{\log\left(1-\frac{1}{K}\right)}+1$, and the definition of $N^\ast$, to ensure that $N^\ast=N$ it is enough to require
\[
\left|\frac{\log \left(1- \hat{p}_t \right) }{\log \left( 1-\frac{1}{K} \right)}-\frac{\log(1-p)}{\log\left(1-\frac{1}{K}\right)}\right|\leq \gamma
\]
for some $\gamma<1/2$, which is equivalent to requiring
\[
\left|\frac{\log \left(\frac{1- \hat{p}_t}{1-p}\right)}{\log \left( 1-\frac{1}{K} \right)}\right|\leq \gamma.
\]
Let $\beta$ denote the actual difference between  $\hat{p}_t $ and $ p $, so that $ \hat{p}_t = p + \beta$. Therefore, the above is equivalent to
%We know that $ N = \frac{\log \left( 1-p \right) }{\log \left( 1- \frac{1}{k} \right)  } + 1$ and that $ 1 -  \left( 1 - \frac{1}{k} \right) ^{N^{\ast} - 1} = \hat{p} = p + \epsilon_1^{\prime} $.\\
%Solving for $ N^\ast $ yields $ N^{\ast} = \frac{\log \left( 1-p-\epsilon_1^{\prime} \right) }{\log \left( 1 - \frac{1}{k} \right)  } + 1 $ which means that $ |N^{\ast} - N | = |\frac{\log \left( \frac{1 - p - \epsilon_1^{\prime}}{1-p } \right) }{\log \left( 1 - \frac{1}{K}  \right) }| $\\
%Given $\gamma > 0 $ we want $ |N^{\ast} - N| = |\frac{\log \left( \frac{1 - p - \epsilon_1^{\prime}}{1-p } \right) }{\log \left( 1 - \frac{1}{K}  \right) }| < \gamma $.\\
\begin{gather*}
-\gamma \leq \frac{\log \left(  \frac{1-p-\beta }{1-p} \right) }{\log \left( 1 - \frac{1}{K} \right) } \leq \gamma \\
\Longleftrightarrow~~ \gamma \log \left( 1 - \frac{1}{K} \right)  \leq \log \left(  \frac{1-p-\beta }{1-p} \right)  \leq -\gamma \log \left( 1 - \frac{1}{K} \right) \\
\Longleftrightarrow~~  \left( 1 - \frac{1}{K} \right) ^{\gamma} \leq \frac{1-p-\beta }{1-p} \leq  \left( 1 - \frac{1}{K} \right) ^{-\gamma}\\
\Longleftrightarrow~~  \left( 1-p \right)  \left( 1-\frac{1}{K} \right) ^{\gamma} \leq 1 - p - \beta \leq  \left( 1-p \right)  \left( 1-\frac{1}{K} \right) ^{-\gamma}\\
\Longleftrightarrow~~ -1+p+ \left( 1-p \right)  \left( 1-\frac{1}{K} \right) ^{ \gamma } \leq -\beta \leq -1+p+ \left( 1-p \right)  \left( 1-\frac{1}{K} \right) ^{ -\gamma }\\
\Longleftrightarrow~~  1-p- \left( 1-p \right)  \left( 1-\frac{1}{K} \right) ^{ -\gamma } \leq \beta \leq 1-p- \left( 1-p \right)  \left( 1-\frac{1}{K} \right) ^{ \gamma } \\
\Longleftrightarrow~~  \left( 1-p \right) \cdot  \left( 1 -  \left( 1-\frac{1}{K} \right) ^{ -\gamma } \right)  \leq \beta \leq  \left( 1-p \right)  \cdot  \left( 1 -  \left( 1-\frac{1}{K} \right) ^{ \gamma }  \right)
\end{gather*}
Therefore, if we can ensure that $|\hat{p}_t-p|\leq \epsilon_1$, where 
\[
\epsilon_1 = \min\left\{ \left| \left( 1-p \right) \cdot  \left( 1 -  \left( 1-\frac{1}{K} \right) ^{ -\gamma } \right) \right|, \left| \left( 1-p \right)  \cdot  \left( 1 -  \left( 1-\frac{1}{K} \right) ^{ \gamma }  \right) \right| \right\}
\]
for some $\gamma<1/2$ (say $0.49$), we get that $N^*=N$ as required. If $t$ is sufficiently large, this can be done using Hoeffding's inequality: $\hat{p}_t$ is an average of $t$ i.i.d. random variables with expectation $p$, hence with probability at least $1-\delta$, $|\hat{p}_t - p | \leq \epsilon_1 $ provided that $t\geq  \frac{\log(2/\delta)}{2\epsilon_1^2}$. 

%Since $ \hat{p}_t $ is an unbiased estimator of $ p $, we can use Hoeffding's inequality and get that $ \forall \epsilon_1>0 $ we have $ \Pr  \left( |\hat{p}_t - p| \geq \epsilon_1 \right)  \leq 2 e^{-2t { \epsilon_1}^2} $. Therefore, w.p. $ \geq 1- 2e^{-2t { \epsilon_1}^2}$ we have $ |\hat{p}_t - p | \leq \epsilon_1 $.

%We note that the $ \epsilon_1 $ we present here uses the true probability of a collision, $ p $. We remind the reader that as seen before: $ p = 1 -  \left( 1 - \frac{1}{K} \right) ^{N-1} $. Thus, $ \epsilon_1 $ is indeed a function of $ \gamma, K, N$.\\

%
%Notice that, according to Hoeffding's inequality, if we choose the number of samples, $T_e$, large enough then with high probability the true difference between $p$ and $\hat{p}$, $\beta$, will be less than $\epsilon_1$, thus the difference between $N^\ast$  and $N$ will be less than $\gamma$.\\

%The case in which every sample round is a collision is a bad event, for which the estimator $ N^\ast $ is undefined. We will use a union bound over the bad events of not estimating $ \hat{p} $ close enough to $ p $ or estimating $ \hat{p}_{T_e} = 1  $ to give the desired result.\\

We now replace the expression of $\epsilon_1$ above, which is a bit unwieldy, with a simpler lower bound (where we also take $\gamma=0.49$). First, plugging in the expression for $p$, we get

%It now remains to 
%Now if we let the number of samples $T_e $ be $\geq \frac{\log \left( \frac{\delta_2}{2}  \right) }{-2{\epsilon_1}^2}$ this implies that $\Pr \left( |\hat{p} - p| \leq \epsilon_1 \right)  \geq 1 - 2\cdot e ^{-2 \cdot T_e \cdot \epsilon_1^2} \geq 1 - \delta_2 $. Since $ p < 1 $ $ \implies \epsilon_1 > 0 \implies T_e \mbox{ is finite}  $.
%\\We conclude that with probability $\geq 1 - \delta_2 $ we have that $ |N^{\ast} - N| \leq \gamma $.\\
%By choosing $ \gamma = 0.49 $ and rounding $ N^{\ast} $ we will obtain the true value of $ N $ w.p. $ 1 - \delta_2 $.\\
%For the sake of clarity, we lower bound $\epsilon_1$:
\begin{gather*}
\epsilon_1 = \min\left\{ \left| \left( \left( 1 - \frac{1}{K} \right) ^{N-1} \cdot  \left( 1 -  \left( 1-\frac{1}{K} \right) ^{ -0.49 } \right) \right) \right|, \left| \left( \left( 1 - \frac{1}{K} \right) ^{N-1} \cdot  \left( 1 -  \left( 1-\frac{1}{K} \right) ^{ 0.49}  \right) \right) \right| \right\}
\end{gather*}
We first lower bound the first expression:
\begin{gather*}
\left| \left( \left( 1 - \frac{1}{K} \right) ^{N-1} \cdot  \left( 1 -  \left( 1-\frac{1}{K} \right) ^{ -0.49 } \right) \right) \right| = -\left( \left( 1 - \frac{1}{K} \right) ^{N-1} \cdot  \left( 1 -  \left( 1-\frac{1}{K} \right) ^{ -0.49 } \right) \right) \\
\geq \left( 1 - \frac{1}{K} \right) ^{K-1} \cdot \left( -\left(1 -  \left( 1-\frac{1}{K} \right) ^{ -0.49 } \right) \right) \geq \frac{1}{\exp(1)} \cdot \left( -\left(1 -  \left( 1-\frac{1}{K} \right) ^{ -0.49 } \right) \right)
\end{gather*}
We use a Taylor expansion to lower bound $ \left(- 1 +  \left( 1-\frac{1}{K} \right) ^{ -0.49 } \right) $: Considering $ f(x) = \left(- 1 +  \left( 1-x \right) ^{ -0.49 } \right) $, the first derivative  is $f^\prime (x) = 0.49 \cdot \left( 1-x \right) ^{ -1.49 }  $ and the second derivative of $f(x)$ is $f^{\prime\prime}(x) = 0.49 \cdot 1.49 \cdot \left( 1-x \right)^{-2.49}$, which is non-negative for any $x\in [0,1]$. Therefore, $f(x) \geq f(0) + f^\prime (0) \cdot x = 0.49 \cdot x$ for any $x\geq [0,1]$, and replacing $x$ with $1/K$ we get that $-\left( 1 -  \left( 1-\frac{1}{K} \right) ^{ -0.49 } \right) \geq \frac{0.49}{K}$ \\

Similarly, we lower bound the second expression:
\begin{gather*}
\left| \left( \left( 1 - \frac{1}{K} \right) ^{N-1} \cdot  \left( 1 -  \left( 1-\frac{1}{K} \right) ^{ 0.49}  \right) \right) \right| = \left( \left( 1 - \frac{1}{K} \right) ^{N-1} \cdot  \left( 1 -  \left( 1-\frac{1}{K} \right) ^{ 0.49}  \right) \right)\\
\geq \left( 1 - \frac{1}{K} \right) ^{K-1} \cdot  \left( 1 -  \left( 1-\frac{1}{K} \right) ^{ 0.49} \right) \geq \frac{1}{\exp(1)} \cdot  \left( 1 -  \left( 1-\frac{1}{K} \right) ^{ 0.49}\right)
\end{gather*}
We use Taylor expansion again to lower bound $\left( 1 -  \left( 1-\frac{1}{K} \right) ^{ 0.49}\right)$. We look at the function: $f(x) = 1 - \left( 1-x \right)^{0.49}$. The first derivative of $f(x)$ is $f^\prime (x) = 0.49 \cdot \left( 1-x \right)^{-0.51}$ and the second derivative of $f(x)$ is $f^{\prime\prime}(x) = 0.49 \cdot 0.51 \cdot \left( 1-x \right)^{-1.51}$. 
Note that $\forall x \in \left[0,1 \right] : f^{\prime\prime}(x) \geq 0$. Thus we get: $f(x) \geq f(0) + f^\prime (0) \cdot x = 0.49 \cdot x$. Thus the lower bound is again:
$\left| \left( \left( 1 - \frac{1}{K} \right) ^{N-1} \cdot  \left( 1 -  \left( 1-\frac{1}{K} \right) ^{ 0.49 } \right) \right) \right| \geq \frac{0.49}{K \cdot \exp(1)}$

Combining the above, we showed that
\begin{gather*}
\epsilon_1 \geq \frac{0.49}{\exp(1) \cdot K } \geq \frac{0.1}{K}.
\end{gather*}
Taking this value for $\epsilon_1$, we get that if we run the learning phase for $\left\lceil \frac{\log\left(\frac{2}{\delta_2 }  \right) }{2{\epsilon_1}^2 } \right\rceil $ rounds, then w.p. $ \geq 1 - \delta_2 $, we have that $ N^{\ast} = N $.
%\end{proof}

\subsubsection{Proof of Lemma~\ref{lemma:fixing} }
%We begin by restating the lemma:\\
%\textbf{Lemma~\ref{lemma:fixing}.}
%\textit{Denote by $ R^F   $ the regret accumulated due to players running the musical chairs subroutine.\\
%\begin{gather*}
%\mathbb{E}[R^F  ] \leq 2 \cdot N^2 \cdot e^2
%\end{gather*}
%}
%\begin{proof}

We remind the reader that the musical chairs phase is when a set of $N$ players who, with high probability, have learned an $\epsilon-$correct ranking each choose an arm uniformly at random from the best $N$ arms and stay 'fixed` on that arm until the end of the epoch or game. Thus once a player has 'fixed`, the only case in which she can contribute regret is if another non-fixed player collides with her.

We will denote by $N$ to be the number of players starting the musical chairs phase. let $r$ be the number of rounds since the start of the musical chairs phase (i.e. when the musical chairs phase starts, $r = 0$).\\

Denote by $ T_f $ the time it takes for one player running the musical chairs subroutine to become fixed. We will first bound $ T_f $.

$N_m$ will denote the maximum number of players (if the game has a dynamic player setting rather than static).

We start by fixing some player who is running the musical chairs subroutine. We will denote by $ v_t $ the number of players that entered late and are not running the musical chairs subroutine, rather they are choosing arms uniformly at random. For any round $ t $ after the musical chairs phase begins the probability for this player to become fixed is at least:
\begin{gather*}
\sum_{\mbox{all unfixed arms}} \frac{1}{N} \cdot \left(1 - \frac{1}{N}\right)^{N-1} \cdot \left(1 - \frac{1}{K} \right)^{v_t} \geq \frac{1}{N}\cdot \left(1 - \frac{1}{N}\right)^{N-1} \cdot \left(1 - \frac{1}{K} \right)^{N_m - N}
\end{gather*}
In the above expression the first term is the probability that the player we are considering chooses the specific arm in the summation. The second term is a lower bound on the probability that all players who are running the musical chairs subroutine and are unfixed do not choose that specific arm since the actual probability is $ (1 - \frac{1}{N})^{N_r} $, where $ N_r $ is the number of player remaining unfixed, which is greater than or equal to $(1 - \frac{1}{N})^N $. The third term is a lower bound on the probability that all players who entered late in the epoch, who choose arms from all $ K $ arms uniformly at random, do not choose the specified arm. In the static setting this term would be 1.

We use the convention that $ (1 - \frac{1}{N})^{N-1} $ equals 1 when $ N $ is 1 since in this case the probability that the specified player becomes fixed is $ 1 $ times the probability that no new player (who is exploring randomly) chooses the last unavailable arm.

We continue bounding the probability that at any round $t$ some player running the musical chairs subroutine, who has not fixed already, becomes fixed:
\begin{gather*}
 \frac{1}{N}\cdot\left(1 - \frac{1}{N}\right)^{N-1} \cdot \left(1 - \frac{1}{K} \right)^{N_m - N} \geq \frac{1}{N}\cdot \left(1 - \frac{1}{N}\right)^{N-1} \cdot \left(1 - \frac{1}{N_m} \right)^{N_m - 1}
\end{gather*}

In the above inequality we use the fact that $N_m \leq K$.\\

Now we use the fact that $(1 - \frac{1}{x})^{x-1} \geq \frac{1}{\exp\left(1\right)}$ for $x\geq 1$ to get that 
\begin{gather*}
\frac{1}{N}\cdot\left(1 - \frac{1}{N}\right)^{N-1} \cdot \left(1 - \frac{1}{N_m} \right)^{N_m - 1} \geq \frac{1}{\exp\left(2\right)} \cdot \frac{1}{N}
\end{gather*}

Now we have established that for any player running the musical chairs subroutine, who has not become `fixed' yet, the probability, at any round $t$, to become fixed in the next round is at least $\frac{1}{\exp\left(2\right)} \cdot \frac{1}{N}$. So for any player, from the moment they begin running the musical chairs subroutine the expected time it takes to become fixed is at most the expected number of times flipping a biased coin with probability $\frac{1}{\exp\left(2\right)} \cdot \frac{1}{N}$, which is $\exp\left(2\right) \cdot N$.

Thus the expected time it takes any player to fix is at most $\exp\left(2\right) \cdot N$.

Notice that after the learning phase in every collision there is at most one fixed player. The regret, at any round after the learning phase, is bounded by two times the number of unfixed players. Therefore the total regret accumulated due to any player running the musical chairs subroutine is bounded by two times the time is takes her to `fix'. Denote by $ T_f^i $ the time it takes player $ i $ to `fix' on an arm. The total expected regret accumulated by players running the musical chairs subroutine is:
$ \leq \mathbb{E}\left[ \sum_{i=1}^N 2 \cdot T_f^i \right] $
Each player running the musical chairs subroutine can only contribute at most 2 to the 

Let $r_{i,t}$ be an indicator variable which equals 1 if player $i$ incurred regret at round $t$ and $0$ otherwise. We will denote by $F\subseteq \left[N\right]$ the set of players who started running the musical chairs subroutine and fixed on an arm and by $U\subseteq \left[N\right]$ the set of players who have not fixed yet.

We will now bound the expected regret due to players running the musical chairs subroutine. Notice that we do not include any regret due to players who entered late and are randomly exploring. 
\begin{gather*}
\leq \mathbb{E}\left[ \sum_{i=1}^N \sum_{t=T_0 +1}^T r_{i,t} \right] \leq \mathbb{E}\left[ \sum_{t=T_0 +1}^T \sum_{i \in U} r_{i,t} + \sum_{j \in F} r_{j,t} \right] \\
\leq  \mathbb{E}\left[ \sum_{t=T_0 +1}^T \sum_{i\in U} r_{i,t} + \sum_{i\in U} r_{i,t} \right] \leq \mathbb{E}\left[ \sum_{t=T_0 +1}^T 2 \cdot \sum_{i \in U} r_{i,t} \right]\\
\leq \mathbb{E}\left[ \sum_{t=T_0 +1}^T 2 \cdot \sum_{i \in U} I_{\mbox{player $i$ not fixed at round $t$}} \right] \leq \mathbb{E}\left[ \sum_{t=T_0 +1}^T 2 \cdot N \cdot I_{\mbox{player $i$ not fixed at round $t$}} \right] \\
 \leq 2 \cdot N \cdot T_f
\end{gather*}

where the third inequality ($\sum_{j \in F} r_{j,t} \leq \sum_{i \in U} r_{i,t}$) is due to the fact that $\sum_{j \in F} r_{j,t}$ is upper bounded by the total number of collisions at round $t$ (since there can only be one fixed player per collision) and the total number of collisions is upper bounded by $ \sum_{i \in U} r_{i,t}$.

In conclusion, we have that the expected regret due to players who run the musical chairs subroutine is $\leq 2 \cdot T_f \cdot N = 2 \cdot \exp\left(2\right) \cdot N^2$

%\end{proof}

\subsection{Analysis of DMC algorithm}

Let $ T_1 $ denote the epoch length, $ T_f = \mathbb{E}[F] \leq \exp\left(2\right) \cdot N_m  $ be the expected time it takes any player to fix on an arm during the `Musical Chairs' period, $ C = T_0 + T_f$, and $ N_m $ be the maximum number of active players at any given round.\\

\subsubsection{Proof of Theorem~\ref{thm:DMC} }
%We begin by restating the theorem:\\
%\textbf{Theorem~\ref{thm:DMC}.}
%\textit{Let $N_m$ be an upper bound on the number of active players at any time point;
%$\Delta_{min}=\min_{i=1,\ldots,N_m}\mu_i-\mu_{i+1}$ the minimal gap between
%the best $N_m+1$ arms, with a known lower bound $\epsilon>0$; and $x$ be an
%upper bound on the totals number of players entering and leaving during $T$ rounds.
%Then with arbitrarily high probability, the expected regret of the DMC
%algorithm (over the rewards), using $\Theta(\sqrt{xT})$ epochs with
%$\tilde{O}(1)$ learning rounds at the beginning of each epoch, is
%\begin{gather*}
%\mathbb{E}[R]\leq \tilde{O} \left( \sqrt{xT}\right)
%\end{gather*}
%where the $ \tilde{O}  $ hides logarithmic factors.}
%\begin{proof}
Using lemma~\ref{lemma:DMC} we will now compute the optimal epoch length, $ T_1 $.\\

We have that the expected regret is
\begin{gather*}
\leq \frac{T}{T_1} \cdot \left( N_m \cdot \left(T_0 + 2 \cdot T_f \right)\right) + e\cdot N_m \left(T_1 - T_0 \right) + l\left(T_1 - T_0 \right)\\
\leq \frac{T}{T_1} \cdot \left( K \cdot \left(T_0 + 2 \cdot T_f \right)\right) + e\cdot K \left(T_1 - T_0 \right) + l\left(T_1 - T_0 \right)
\end{gather*}
We differentiate with respect to $ T_1 $ and set equal to zero to find the optimal value of $ T_1 $ (up to rounding): \\
$T_1 = \left \lceil \sqrt{ \frac{T \cdot K \cdot \left(T_0 + 2 \cdot T_f \right)}{K \cdot e + l} } \right \rceil$ and since $e+l \leq x$ we take $T_1 = \left \lceil \sqrt{ \frac{T \cdot \left(T_0 + 2\cdot T_f \right)}{ x  } } \right \rceil$
\\
Note that we require $ T_1 $ to be greater than $ T_0 $ (the epoch length must be at least as long as the learning period ) . Thus if the optimal $ T_1 $ is less than $ T_0 $, we set $ T_1 = T_0 $.\\
Plugging $ T_1 $ into the regret bounds yields that the expected regret is $\leq O \left( \frac{T}{T_1} \cdot T_0 + x \cdot T_1  \right) $ where \\
 $ T_1 = O \left(  \sqrt{\frac{T \log \left( T \right)}{x} }\right) $ and $ T_0 = O \left( \log \left(  T \right)  \right)  $.
\\Thus we get that the expected regret is $\leq O \left(   \frac{T}{\sqrt{\frac{T \log \left( T \right)}{x} }} \cdot \log(T) + x \cdot \sqrt{\frac{T \log \left( T \right)}{x} }  \right)\\ = O\left( \sqrt{\frac{T \cdot x }{\log(T)}} \cdot \log(T) + x \cdot \sqrt{\frac{T \log(T)}{x}} \right) = O\left( \sqrt{T x} \cdot \frac{\log(T)}{\sqrt{\log(T)}} + \sqrt{T x} \cdot \sqrt{\log(T)} \right)\\
= \tilde{O}\left( \sqrt{T x} \right) $ where the $ \tilde{O}  $ hides logarithmic factors in $x,T$, and $x$ is an upper bound on the number of players entering and exiting.
%\end{proof}

\subsubsection{Proof of Lemma~\ref{lemma:DMC} }
%We begin by restating the lemma:\\
%\textbf{Lemma~\ref{lemma:DMC}.}
%\textit{$  \forall $  $  0< \delta < 1 $ and $ 0 <\epsilon < \Delta_{min} $, w.p. $ \geq 1 - \delta $ the expected regret of the Dynamic MC algorithm played for $ T $ rounds, with parameters: \\
%$ T_0 = \left\lceil \max\left( \frac{K}{2} \cdot \ln \left( \frac{2 \cdot K^2}{\frac{\delta}{2 \cdot T}} \right), \frac{16 \cdot K}{ \epsilon^2 } \cdot \ln\left( \frac{4 \cdot K^2}{\frac{\delta}{2 \cdot T}} \right) , 50 \cdot \log\left(\frac{2}{1 - \frac{\delta}{2 \cdot T}}  \right)  \right) \right\rceil $ and $ T_1$ chosen such that $ T_1 > T_0 $, is: \\
%}
%
%
%$ \leq \frac{T}{T_1} \cdot \left( N_m \cdot \left(T_0 + 2 \cdot T_f \right)\right) + e\cdot 2 \left(T_1 - T_0 \right) + l\left(T_1 - T_0 \right)  $ \\
%
%\textit{
%where $e$ is the total number of players entering, $l$ is the total number of players leaving, $T_f$ is the expected time for any player to fix on an arm, $T_f \leq N_m \cdot e^2$, and $\Delta_{min} $ is $\min_{i} \mu_i - \mu_{i+1}$ where $\mu_i$ is the expected reward of the $i^{th}$ best arm. \\
%}

%\begin{proof}
We start by noting that the number of epochs is at most $ \lceil \frac{T}{T_1} \rceil$, and we compute a bound on the expected regret per epoch and sum over the epochs using the linearity of expectation.\\
The regret per epoch is composed of three terms:
\\
\begin{itemize}
\item Regret due to learning and fixing on an arm
\item Regret due to entering players
\item Regret due to leaving players
\end{itemize}
We will now compute each of these terms.
\\
\paragraph*{Regret due to learning and fixing on an arm}
For each epoch, we have at most $N_m$ players who are learning for $T_0$ rounds and fixing on an arm, each one taking $ T_f $ rounds, in expectation. For every given round of learning or fixing an upper bound on the expected regret is $ N_m $ since the best expected reward minus the worst expected reward is less than 1. Also, as shown in lemma~\ref{lemma:fixing}, the expected regret due to players who run the musical chairs subroutine is at most $2 \cdot \exp\left(2\right) \cdot N_m^2  = 2 \cdot T_f \cdot N_m $. This means that the total expected regret for this term, per epoch, is $ \leq   N_m \cdot \left( T_0 + 2 \cdot T_f \right) $. \\
\paragraph*{Regret due to entering players}
We remind the reader that players cannot enter during the learning period. Since a newly entered player learns until the end of the epoch she contributes at most $ T_1 - T_0 $ regret due to learning. During this time period the player may also collide with other players who are already fixed or collide with players who have not finished fixing after the expected time it takes to fix. A newly entering player can collide with at most $1$ other fixed player. Since we already count the regret of a any round of a non-fixed player as regret we only need to add regret added due to collisions with fixed players. Thus the total regret a single newly entering player can contribute is $2 \cdot \left(T_1 - T_0 \right)$ and the total regret due to all newly entering players is at most $e_i \cdot 2 \cdot \left( T_1 - T_0 \right)$ per epoch, where $e_i$ is the number of players who enter at epoch $i$.\\
\paragraph*{Regret due to leaving players}
We remind the reader that players cannot leave during the learning period. \\
Every player that leaves can contribute at most $ 1 $ regret at each round, since in the worst case this player causes the best arm to remain unused during the fixed period. This would mean that for every player that leaves, per epoch, we have at most $ T_1 - T_0 $ added regret. Denote by $l_i$ the number of players that leave during epoch $i$ ($\sum_{i=1}^{\frac{T}{T_1}} l_i = l$). For all players that leave in epoch $i$, this amounts to having an added regret of $  l_i \cdot \left( T_1 - T_0  \right) $. \\
\\

Now summing over the all epochs we get that the total expected regret of the DMC algorithm is:
\begin{gather*}
\leq \sum_{i=1}^{\frac{T}{T_1}} \left( N_m \cdot \left(T_0 + 2T_f \right) + e_i\cdot N_m \left(T_1 - T_0  \right) + l_i \left(T_1 - T_0 \right) \right)\\
 = \frac{T}{T_1} \cdot \left( N_m \cdot \left(T_0 + 2T_f \right)\right) + e\cdot N_m \cdot \left(T_1 - T_0  \right) + l\left(T_1 - T_0 \right)
\end{gather*}

For the regret bound to be correct, we need to ensure that players learn an $\epsilon$-correct ranking and correctly estimate the number of players at every epoch. By using lemma~\ref{lemma:mclearn} and lemma~\ref{lemma:unknown} with confidence parameters set to $ \frac{\delta}{2 \cdot T} $, and taking the union bound over all epochs, we ensure that with high probability the players learn the true rankings and estimate the number of players correctly at each epoch, and thus we get that w.p. $\geq 1 - \delta $ we have that the regret bound holds. For this reason $ T_0 $ includes a $ \log\left(T \right)$ factor, as stated above.\\

Notice that letting $ T_0 = \left\lceil \max\left( \frac{K}{2} \cdot \ln \left( \frac{2 \cdot N_m \cdot K}{\frac{\delta}{2 \cdot T}} \right), \frac{16 \cdot K}{ \epsilon^2 } \cdot \ln\left( \frac{4 \cdot N_m \cdot K}{\frac{\delta}{2 \cdot T}} \right) , 50 \cdot \log\left(\frac{2}{1 - \frac{\delta}{2 \cdot T}}  \right)  \right) \right\rceil $ is only possible if one knows $N_m$. If $N_m$ is unknown, then one can increase $T_0$ and set it to $ T_0 = \left\lceil \max\left( \frac{K}{2} \cdot \ln \left( \frac{2 \cdot K^2}{\frac{\delta}{2 \cdot T}} \right), \frac{16 \cdot K}{ \epsilon^2 } \cdot \ln\left( \frac{4 \cdot K^2}{\frac{\delta}{2 \cdot T}} \right) , 50 \cdot \log\left(\frac{2}{1 - \frac{\delta}{2 \cdot T}}  \right)  \right) \right\rceil $ and the lemma holds since we assume that $N_m < K$.

%\end{proof}

\subsection{Analysis of scenarios from section~\ref{comparison}}

\subsubsection{Proof of Theorem~\ref{thm:discuss2} }

We will denote by $ p_i $ player $ i $ for $ i\in\{1,2\} $ and the highest ranked arm as $ a_1 $. In a nutshell, the argument goes as follows:
We show that the probability that $ p_2 $ does not ever learn the ranking of $ a_1 $  in rounds $ [\left\lceil \frac{T}{2}\right\rceil, \left\lceil \frac{T}{2} + f \cdot T \right\rceil] $, and that $ p_1 $ stayed on arm $a_1$ during rounds $ [\left\lfloor \frac{T}{4}\right\rfloor, \left\lceil \frac{T}{2}\right\rceil]$, is at least some constant $ b $, not dependent on $ T $. As a result, from time $\left\lceil \frac{T}{2} + f \cdot T \right\rceil$ onwards, the exploration probability of $ p_2 $ will be at least $ \epsilon_2  \left( f \cdot T \right)  = \frac{c \cdot K^2}{d^2 \cdot  \left( K-1 \right)  \cdot f \cdot T} $  ( this is because $ p_2 $ has played for at least $ f \cdot T $ rounds ) . Thus the expected time for $ p_2 $ to explore $ a_1 $, given that she does not rank $ a_1 $ highly enough to exploit it, is $ \geq \frac{1}{\epsilon_2  \left( f \cdot T \right)  } = \Omega \left( T \right)  $. So the expected regret will be $ \geq b \cdot \Omega \left( T \right)  +  \left( 1-b \right)  \cdot 0 = \Omega \left( T \right) $.

Denote by $A$ the event that $p_2$ never learns $a_1$ in rounds $[\lfloor\frac{T}{2}\rfloor, \lceil\frac{T}{2} + fT\rceil $.
Denote by $B$ the event that $p_1$ does not leave $a_1$ in rounds $[\lfloor \frac{T}{4}\rfloor, \lceil\frac{T}{2}\rceil ]$. Since $\Pr \left( A \bigwedge B \right) = \Pr \left(A | B \right) \cdot \Pr \left( B\right)$ we will lower bound both of these terms, $\Pr \left(A | B \right)$ and $\Pr \left( B \right)$. We start by lower bounding $\Pr \left( B \right)$:
\begin{gather*}
\Pr \left( p_1 \mbox{ leaves } a_1 \mbox{at least once during the rounds } \lfloor \frac{T}{4}\rfloor, \lceil \frac{T}{2} \rceil \right) \leq\\
\sum_{t= \lfloor \frac{T}{4} \rfloor}^{\lceil \frac{T}{2} \rceil} \Pr \left( p_1 \mbox{ explores at round } t \right) = \sum_{t= \lfloor \frac{T}{4} \rfloor}^{\lceil \frac{T}{2} \rceil} \frac{c K^2}{d^2 (K-1) t}\\
\leq \sum_{t= \lfloor \frac{T}{4} \rfloor}^{\lceil \frac{T}{2} \rceil} \frac{c K^2}{d^2 (K-1) \lfloor\frac{T}{4}\rfloor} \leq \left(\lfloor \frac{T}{4} \rfloor \right) \frac{c K^2}{d^2 (K-1) \lfloor\frac{T}{4}\rfloor} = \frac{c K^2}{d^2 (K-1)}
\end{gather*}
Thus $\Pr \left( B \right) \geq  1 - \frac{c K^2}{d^2 (K-1)}$.

We now compute a lower bound on $ \Pr\left( A | B \right) $, the probability that $ p_2 $ does not learn the ranking of $ a_1 $  in rounds $ [\left\lceil \frac{T}{2}\right\rceil, \left\lceil \frac{T}{2} + f \cdot T \right\rceil] $ given that $p_1 $ never left $a_1$ during the rounds $[\lfloor \frac{T}{4}\rfloor, \lceil\frac{T}{2}\rceil ]$ :\\
\begin{gather*}
 \Pr\left(\overline{A}|B\right) = \Pr \left( \mbox{$ p_2 $ learns $ a_1 $ in rounds $[\left\lceil \frac{T}{2}\right\rceil, \left\lceil \frac{T}{2} + f \cdot T \right\rceil]  $}|B \right) \\
 \leq \Pr \left( \mbox{$ p_2 $ has one successful sample of $ a_1 $} |B \right) \\
 = \Pr \left( \mbox{$ p_2 $ samples $ a_1 $ and $ p_1 $ is absent} |B \right) \\
 \leq \Pr \left( \mbox{$ p_1 $ is absent at some round in $[\left\lceil \frac{T}{2}\right\rceil, \left\lceil \frac{T}{2} + f \cdot T \right\rceil]$}|B \right) \\
 = \sum_{n= \frac{T}{2} }^{\frac{T}{2} + f \cdot T} \Pr \left( \mbox{$ p_1 $ left $ a_1 $ for the first time at round n, and any number of times afterwards}|B \right)
\end{gather*}

Denote by $ A_n $ the event that $ p_1 $ leaves $ a_1 $ for the first time since round $ \lceil \frac{T}{2} \rceil $ at round $ n $ and by $B_n$ the event that $p_1$ has not left $a_1$ from round $ \lceil \frac{T}{2} \rceil $ until round $n-1$.\\
So we get:

\begin{gather*}
 \Pr \left( \mbox{$ p_2 $ learns $ a_1 $ in rounds $[\left\lceil \frac{T}{2}\right\rceil, \left\lceil \frac{T}{2} + f \cdot T \right\rceil] $}|B \right)  \leq \sum_{n= \lceil \frac{T}{2} \rceil }^{\lceil \frac{T}{2} + f \cdot T \rceil} \Pr \left(  A_n |B \right) \\
 = \sum_{n= \lceil \frac{T}{2} \rceil }^{\lceil \frac{T}{2} + f \cdot T \rceil} \Pr \left( \mbox{$ p_1 $ leaves due to exploration at round $ n $}| B_n, B  \right) \\ \cdot \Pr \left( B_n |B \right)  + \Pr \left( \mbox{$ p_1 $ leaves due to a collision at round $ n $}| B_n, B  \right) \cdot \Pr \left(B_n |B  \right)   \\
\leq  \sum_{n= \lceil \frac{T}{2} \rceil }^{\lceil \frac{T}{2} + f \cdot T \rceil} \Pr \left( \mbox{$ p_1 $ leaves due to exploration at round $ n $}| B_n, B  \right)  \\+ \Pr \left( \mbox{$ p_1 $ leaves due to a collision at round $ n $}| B_n, B  \right) \\
\underset{\ast}{\underbrace{\leq}} \sum_{n= \lceil \frac{T}{2} \rceil }^{\lceil \frac{T}{2} + f \cdot T \rceil} \frac{c \cdot K^2 }{d^2 \cdot  \left( K-1 \right)  \cdot n} +  \left( 1 -  \left( 1 - \alpha^{\lfloor \frac{T}{4} \rfloor} \right)  \right) \\
= \sum_{n= \lceil \frac{T}{2} \rceil }^{\lceil \frac{T}{2} + f \cdot T \rceil} \frac{c \cdot K^2 }{d^2 \cdot  \left( K-1 \right)  \cdot n} + \alpha^{\lfloor \frac{T}{4} \rfloor } \leq \sum_{n= \lceil \frac{T}{2} \rceil }^{\lceil \frac{T}{2} + f \cdot T \rceil} \frac{c \cdot K^2 }{d^2 \cdot  \left( K-1 \right)  \cdot  \frac{T}{2} } + \alpha^{\lfloor \frac{T}{4} \rfloor } \\ =  g \cdot \left( \frac{T}{2} + f\cdot T - \frac{T}{2} \right)  \cdot  \left( \frac{c \cdot K^2 }{d^2 \cdot  \left( K-1 \right)  \cdot \frac{T}{2}	} + \alpha^{\lfloor \frac{T}{4} \rfloor }  \right) \\
= g \cdot \frac{c \cdot K^2 \cdot 2 \cdot f }{d^2 \cdot  \left( K-1 \right) } + g \cdot f \cdot T \cdot \alpha^{\lfloor \frac{T}{4} \rfloor}
\end{gather*}
$g$ is a constant that when multiplied by $\left( \frac{T}{2} + f\cdot T - \frac{T}{2} \right) $ ensures that these numbers are whole numbers.\\
The inequality labeled $ \ast $ is due to the fact that since it is given that $ p_1 $ always exploited arm $ 1 $ from rounds $ \lfloor \frac{T}{4} \rfloor $ to $ \lceil \frac{T}{2} \rceil $ then she would have a persistence probability $  \geq 1-\alpha^{\lfloor \frac{T}{4} \rfloor} $.\\
This fact can be seen from the following calculation:
\begin{gather*}
p_{t+1} = p_t \cdot \alpha +  \left( 1 -\alpha \right)  =  \left( p_{t-1}\cdot \alpha +  \left( 1 - \alpha \right)  \right)  \cdot \alpha +  \left( 1- \alpha \right) \\
= p_{t-1} \cdot \alpha^2 +  \left( 1- \alpha \right)  \left( \alpha + 1 \right)  =  \left( p_{t-2} \cdot \alpha +  \left( 1 - \alpha  \right)   \right)  \cdot \alpha^2 +  \left( 1-\alpha \right)  \left( \alpha + 1 \right) \\
= p_{t-2} \cdot \alpha^3 +  \left( 1- \alpha \right)  \cdot  \left( \alpha^2 + \alpha + 1 \right)
\end{gather*}
Which implies that if $p_1$ did not leave $a_1$ from round $\lfloor \frac{T}{4} \rfloor$ until round $\lceil \frac{T}{2} \rceil$ then the persistence parameter of $p_1$ at round $\lceil \frac{T}{2} \rceil$ will be $p_{\lceil \frac{T}{2} \rceil} = p_{\lfloor \frac{T}{4} \rfloor} \cdot \alpha^{\lfloor \frac{T}{4} \rfloor} +  \left( 1 - \alpha \right)  \cdot  \left( 1 + \alpha + \alpha^2 + ... + \alpha^{\lfloor \frac{T}{4} \rfloor - 1} \right) = p_{\lfloor \frac{T}{4} \rfloor} \cdot \alpha^{\lfloor \frac{T}{4} \rfloor} +  \left( 1 - \alpha \right)  \cdot \sum_{r=0}^{\lfloor \frac{T}{4} \rfloor - 1} \alpha^r = p_{\lfloor \frac{T}{4} \rfloor} \cdot \alpha^{\lfloor \frac{T}{4} \rfloor} +  \left( 1 - \alpha \right)  \cdot \frac{1-\alpha^{\lfloor \frac{T}{4} \rfloor} }{1 - \alpha} \geq 1-\alpha^{\lfloor \frac{T}{4} \rfloor} $\\

Thus we see that the probability that $ p_2 $ does not learn $ a_1 $ in rounds $[\left\lceil \frac{T}{2}\right\rceil, \left\lceil \frac{T}{2} + f \cdot T \right\rceil]  $ given that $p_1$ did not leave $a_1$ in rounds $[\lfloor \frac{T}{4} \rfloor, \lceil \frac{T}{2} \rceil]$ is  $ \geq 1 -  \left( \frac{c \cdot K^2 \cdot 2 \cdot f }{d^2 \cdot  \left( K-1 \right) } + f\cdot T \cdot \alpha^{\lfloor \frac{T}{4} \rfloor} \right)  $.

We will now show that this is lower bounded by some constant, not dependent on $ T $, by showing that:\\
$ \Pr \left( \mbox{$ p_2 $ learns $ a_1 $ in rounds $[\left\lceil \frac{T}{2}\right\rceil, \left\lceil \frac{T}{2} + f \cdot T \right\rceil]  $} \right)  \leq \frac{1}{2}$ and thus:\\
$ \Pr \left( \mbox{$ p_2 $ does not learn $ a_1 $ in rounds $[\left\lceil \frac{T}{2}\right\rceil, \left\lceil \frac{T}{2} + f \cdot T \right\rceil]  $} \right)  \geq \frac{1}{2}$

We will now show that either the following two statements are true, in which case we have shown that the probability of $ p_2 $ not learning $ a_1 $ is greater than some constant, or they are not both true and the MEGA algorithm will have linear regret due to other reasons, inherent to the algorithm:\\
\begin{itemize}
\item $ \frac{c \cdot K^2 \cdot 2 \cdot f }{d^2 \cdot  \left( K-1 \right) } \leq \frac{1}{4} $
\item $ f\cdot T \cdot \alpha^{\frac{T}{4}} \leq \frac{1}{4} $
\end{itemize}

These two statements depend on the choice of parameters of MEGA: $ c,d, \alpha $. For the first statement we show that if these parameters are chosen in a way which contradicts the inequality, then the MEGA algorithm results in linear regret, thus satisfying the main claim in any case. The second statement is true by assumption.\\
We start with the first statement:\\

$ \frac{c \cdot K^2 \cdot 2 \cdot f }{d^2 \cdot  \left( K-1 \right) } \leq \frac{1}{4} $\\
For this to be true we need: $ f \leq \frac{d^2 \cdot  \left( K-1 \right) }{8 \cdot c \cdot K^2} $. At the same time we want f large enough such that $ f \cdot T \geq X $ for some constant $ X \geq \frac{c \cdot K^2 }{d^2 \cdot  \left( K-1 \right) } $ to ensure that the exploration coefficient for $ p_2 $ will be less than $ 1 $ when $ p_1 $ leaves. \\
If we have an $ f $ such that: $ \frac{X}{T} \leq f \leq \frac{d^2 \cdot  \left( K-1 \right) }{8 \cdot c \cdot K^2}$ then we are done. If we cannot find such an f then we are in the case where:\\
$ \frac{d^2 \cdot  \left( K-1 \right) }{8 \cdot c \cdot K^2} \leq \frac{X}{T} $. \\
This happens if: $ d^2 \leq \frac{8 \cdot c \cdot K^2 \cdot X}{ \left( K-1 \right)  \cdot T} $ or $ c \geq \frac{T \cdot d^2 \cdot  \left( K-1 \right) }{8 \cdot K^2 \cdot X} $. \\
We note that on those two cases the regret will be linear due to exploration, since:\\
If $ d^2 \leq \frac{8 \cdot c \cdot K^2 \cdot X}{ \left( K-1 \right)  \cdot T} $ then $ \forall t \leq \frac{T}{8 \cdot X} : \epsilon_t = \frac{c \cdot K^2}{d^2 \cdot  \left( K-1 \right)  \cdot t} \geq \frac{c \cdot K^2}{\frac{X \cdot 8 \cdot c \cdot K^2}{ \left( K-1 \right)  \cdot T} \cdot  \left( K-1 \right)  \cdot t} = \frac{T}{8 \cdot X \cdot t}  \geq \frac{T}{8 \cdot X \cdot \frac{T}{8 \cdot X}} = 1 $. \\

If $ c \geq \frac{T \cdot d^2 \cdot  \left( K-1 \right) }{8 \cdot K^2 \cdot X} $ then $ \forall t \leq \frac{T}{8 \cdot X} : \epsilon_t = \frac{c \cdot K^2}{d^2 \cdot  \left( K-1 \right)  \cdot t} \geq \frac{\frac{T \cdot d^2 \cdot  \left( K-1 \right) }{8 \cdot K^2 \cdot X} \cdot K^2}{d^2 \cdot  \left( K-1 \right)  \cdot t} = \frac{T}{8 \cdot X \cdot t} \geq \frac{T}{8 \cdot X \cdot \frac{T}{8 \cdot X}} = 1 $. \\

Thus, in both cases, players will be exploring for at least $ \frac{T}{8 \cdot X} $ rounds, resulting in linear regret. \\

We now consider the second statement, i.e. : $ f\cdot T \cdot \alpha^{\frac{T}{4}} \leq \frac{1}{4} $\\
Note that for this to be true, we need: $ \alpha^\frac{T}{4} \leq \frac{1}{4 \cdot f \cdot T} \implies \alpha \leq  \left( \frac{1}{4 \cdot f \cdot T} \right) ^\frac{4}{T} $. This value is very close to $ 1 $, and approaches $ 1 $ as $ T \rightarrow \infty $. Thus for any reasonable choice of $ \alpha $, the expected regret is $ \Omega \left( T \right)  $ and when $ \alpha $ is not in this range, i.e. very close to $ 1 $, then $ p $ will hardly deviate from $ p_0 $ which means that during a collision event both players have the same chance of not persisting and making an arm unavailable. This will make the unavailability mechanism not functional, and moreover, it will cause more collisions by forcing the players onto the second arm. For clarity we use a weaker bound on $ \alpha $: $ \alpha \leq 1 - \frac{4\log(4fT)}{T} $. We show that this is weaker:
\begin{gather*}
\frac{1}{4fT}^{\frac{4}{T}} = \exp\left(\log(\frac{1}{4fT}) \cdot \frac{4}{T}\right) = \exp\left(-\log(4fT) \cdot \frac{4}{T}\right) \geq 1 - \log(4fT) \cdot \frac{4}{T}
\end{gather*}
Thus, by assumption, the second statement is true as well.
\\
Thus we have shown that $\Pr\left( A \bigwedge B \right) \geq \frac{1}{2 } \cdot \left( 1 - \frac{c K^2}{d^2 (K-1)}\right)$. So the expected regret of the MEGA algorithm is $\Omega(T)$ w.p. $\geq \frac{1}{2 } \cdot \left( 1 - \frac{c K^2}{d^2 (K-1)}\right)$. This is true since it would take $\Omega\left(T\right)$ rounds, in expectation, for player 2 to learn the true ranking of arm 1 after player 1 has left the game.  The upper bound for the DMC algorithm calculated in the dynamic setting is also an upper bound for this scenario taking $ x = 2 $ which yields an expected regret $ \leq O \left( \sqrt{T} \right)  $.

%\end{proof}

\subsubsection{Proof of Theorem~\ref{thm:discuss1} }
%We begin by restating the Theorem:\\
%\textbf{Theorem~\ref{thm:discuss1}.}
%\textit{
%In the multi-player MAB setting with one player leaving every $(2 \cdot r) \cdot
%\lceil T^\lambda \rceil$ rounds, and one player entering every $(2 \cdot r + 1)
%\cdot \lceil T^\lambda\rceil $ rounds, for $r = 0,1,2,...$, and  $\lambda > \beta$,
%we have that
%\begin{enumerate}
%\item The regret upper bound of the MEGA algorithm is at least
%    $O \left( T^{1 - \left( \lambda - \beta \right)} \right)$
%\item The expected regret upper bound of the DMC algorithm is
%    $\tilde{O} \left( T^{1- \frac{\lambda}{2}} \right)$
%\end{enumerate}
%}
%\begin{proof}

%In this simple scenario players alternate between exiting and entering in consecutive time points of $ \lceil T^\lambda \rceil $. This ensures that the number of players remains above 0 and below $K$. We compare the upper bound, using the analysis given in \cite{avner2014concurrent}, to an upper bound of the DMC algorithm.\\

Since one player leaves every $ X = (2 \cdot r) \cdot  \lceil T^\lambda \rceil $ rounds, the rounds at which a player leaves are: $ 0 \cdot  \lceil T^\lambda \rceil, 2\cdot \lceil T^\lambda \rceil , 4 \cdot  \lceil T^\lambda \rceil,..., \left(\lceil \frac{T}{  \lceil T^\lambda \rceil}\rceil - 1 \right) \cdot \lceil T^\lambda \rceil $. \cite{avner2014concurrent} present a case in which only one player leaves at round $ t $, and causes an added regret of $ t^\beta $. They show that for the scenario of one player leaving, this is a worst case regret bound. We sum over half the total rounds to avoid concerns of counting regret of the last round. Using this bound of $t^\beta$ we get that the total added regret due to all of the players who left is at least:
\begin{gather*}
\sum_{r=0}^{\frac{1}{2} \cdot \lceil \frac{T}{2 \cdot T^\lambda} \rceil } \left( 2 \cdot r \cdot \lceil T^\lambda \rceil  \right) ^\beta \geq\sum_{r=0}^{\frac{1}{2} \cdot \frac{T}{2 \cdot T^\lambda}} \left( 2 \cdot r \cdot T^\lambda  \right) ^\beta = T^{\lambda \cdot \beta } \cdot 2^\beta \cdot \sum_{r=0}^{\frac{1}{2} \cdot \frac{T}{2 \cdot T^\lambda}} r^\beta \\
\geq T^{\lambda \cdot \beta } \cdot 2^\beta \cdot \int_{r=0}^{\frac{1}{2} \cdot \frac{T}{2 \cdot T^\lambda}} r^\beta dr
= T^{\lambda \cdot \beta } \cdot 2^\beta \cdot \frac{1}{\beta + 1 } \cdot \frac{1}{4^{\beta+1}} \cdot \frac{T^{\beta +1}}{T^{\lambda \cdot  \left( \beta + 1 \right) }} \\
=\frac{2^\beta}{\left(\beta + 1\right) \cdot 4^{\beta+1}} \cdot T^{\lambda \cdot \beta + \beta + 1 - \lambda \cdot \beta - \lambda} = \frac{1}{\left(\beta + 1\right) \cdot 2^{\beta + 2}} \cdot T^{1 -  \left( \lambda - \beta  \right) }
\end{gather*}
So we get that:
\begin{gather*}
R_{mega} \left(T \right)  \geq \frac{1}{\left(\beta + 1\right) \cdot 2^{\beta + 2}} \cdot T^{1 -  \left( \lambda - \beta  \right) }
\end{gather*}
Thus the expected regret of the MEGA algorithm in this scenario is at least $ \min \{T,   \frac{1}{\left(\beta + 1\right) \cdot 2^{\beta + 2}} \cdot T^{1 -  \left( \lambda - \beta  \right) } \} $ since regret cannot be larger than linear. Notice that this bound does not include any regret due to learning, collisions, and other sources of regret.\\
We will compare this bound with an upper bound of the DMC algorithm for this scenario.\\
In the analysis of the DMC algorithm in the dynamic setting we calculated regret when at most $x$ players enter or leave. In this case we have $\frac{T}{T^\lambda}$ players leaving or entering. Thus the regret bound is $O\left(\sqrt{xT}\right) = O\left(\sqrt{T \frac{T}{T^\lambda	}}\right) =O\left( T^{1 - \frac{\lambda}{2}}\right)$.

%\end{proof}

\end{document}